%% file: main.tex
\documentclass{article}
\usepackage[T1]{fontenc}
\usepackage{iclr2027_conference,times}
\usepackage{graphicx}
\usepackage{float}
\usepackage{booktabs}
\usepackage{tcolorbox}
\usepackage{algorithm}
\usepackage{algpseudocode}
\usepackage{placeins}
\usepackage{array}
\usepackage{multirow}
\usepackage{fvextra}
\definecolor{RoyalBlue}{RGB}{65,105,225}

\input{math_commands.tex}

\usepackage[hidelinks]{hyperref}
\usepackage{url}
\hypersetup{
  pdftitle={PersonaFactor: Preference Factors with Dynamic Selection for LLM Personalization},
  pdfauthor={Hang Su, Zequn Liu, Jianghan Chao, Qiuyu Tian, Xuesong Lu, Yingce Xia}
}

\title{PersonaFactor: Preference Factors with \\
Dynamic Selection for LLM Personalization}
\author{
  Hang Su\textsuperscript{1,2} \quad Zequn Liu\textsuperscript{2,*} \quad
  Jianghan Chao\textsuperscript{2,3} \quad Qiuyu Tian\textsuperscript{2,4} \\[0.35ex]
  \bfseries Xuesong Lu\textsuperscript{1,*} \quad Yingce Xia\textsuperscript{2,*} \\[1.3ex]
  \small \textsuperscript{1}East China Normal University \qquad
  \textsuperscript{2}Beijing Zhongguancun Academy \\[0.35ex]
  \small \textsuperscript{3}Renmin University of China \qquad
  \textsuperscript{4}Southeast University
}

\iclrshowreviewauthors

\begin{document}

\raggedbottom

\maketitle
\begingroup
\renewcommand{\thefootnote}{\fnsymbol{footnote}}
\footnotetext[1]{\raggedright\hangindent=1.8em\hangafter=1 Corresponding authors: Yingce Xia, Xuesong Lu, Zequn Liu.\newline Contact: \texttt{\{s-sh25, liuzequn, s-cjh25, s-tqy24, xiayingce\}@bza.edu.cn}, \texttt{xslu@dase.ecnu.edu.cn}.}
\endgroup

\begin{abstract}
Although users may express different personalization needs across tasks, these
task-specific behaviors can reflect shared underlying preference mechanisms.
Existing LLM personalization methods often focus on surface-level interactions
or undifferentiated histories, overlooking the shared mechanisms underlying
these behaviors. We introduce \emph{PersonaFactors}, a framework that distills
reusable and interpretable personalization factors from heterogeneous user
interactions. We construct a library of thirteen PersonaFactors by extracting
evidence-grounded rationales and clustering recurring patterns across tasks,
and develop a dynamic factor-guided generation framework that selects factors
relevant to each user's history and current request. Across four personalization
benchmarks and four backbone models, Dynamic PersonaFactor achieves the best
score on every reported metric. Representation analyses further show that
PersonaFactors and user history induce partially overlapping low-dimensional
representations, and that their joint use yields complementary shifts. Taken
together, these findings suggest that factor-based and history-based
personalization can mutually reinforce each other.
\end{abstract}

\input{1_Intro_yinxia.tex}

\input{2_Related_Work.tex}
\input{3_Method_yinxia.tex} 
\input{4_Experiments.tex}
\input{5_Conclusion.tex}
\input{11_AI_Use_Statement.tex}
\input{12_Reproducibility_Statement.tex}

\input{7_Ethics_Statement.tex}

\bibliography{reference}
\bibliographystyle{iclr2027_conference}

\clearpage
\appendix
\input{9_Appendix.tex}
\input{6_Limitations.tex}

\end{document}

%% file: math_commands.tex
\usepackage{amsmath,amsfonts,bm}

\def\eqref#1{equation~\ref{#1}}

\def\1{\bm{1}}

\DeclareMathAlphabet{\mathsfit}{\encodingdefault}{\sfdefault}{m}{sl}
\SetMathAlphabet{\mathsfit}{bold}{\encodingdefault}{\sfdefault}{bx}{n}



%% file: 1_Intro_yinxia.tex
\section{Introduction}
Large language models (LLMs) are increasingly used for personalized applications such as question answering~\citep{su2025personalized,salemi2025lamp}, content creation~\citep{salemi2024lamp,wang2026think}, and task assistance~\citep{cai2025large,wan2025enhancing}. These applications require adapting both the content and presentation of responses to individual users, whose knowledge, goals, constraints, and preferences can lead to different expectations for similar requests.

Existing personalization methods broadly follow two strategies. \textbf{Model adaptation} learns user preferences from historical interactions and incorporates them into the model through parameter adaptation, user representations, activation steering, or decoding control~\citep{zhang2025proper,choi2025copl,qiu2025latent, zhang2025personalized}. These methods provide direct control over generation but typically require additional training or access to model internals.
\textbf{Context conditioning} keeps the underlying model unchanged and represents user preferences as text. It retrieves relevant past interactions or summarizes them into user profiles and memories~\citep{su2025personalized,pan2025secom,zhang2025prime}, then includes this information in the prompt to guide personalized generation.

There are two key challenges in turning user history into effective personalization guidance. \textbf{First, identifying reusable personalization mechanisms.} Historical interactions are highly task-specific, while the same underlying preference may appear differently across scenarios. We therefore need to abstract shared mechanisms beyond surface behaviors. \textbf{Second, selecting the right mechanisms for each request.} Not every user preference is relevant to every query, and using all factors may introduce redundant or conflicting guidance. The challenge is to identify the subset that is most relevant to the current task and most influential for generation.

We introduce \emph{PersonaFactors}, a framework that induces reusable personalization factors from heterogeneous user interactions through evidence-grounded rationales. Specifically, we first use an LLM to extract rationales explaining why a response fits a particular user, cluster recurring rationales across tasks, and canonicalize them into 13 interpretable factors spanning cognitive, contextual, identity, and behavioral dimensions. Given a new user history and query, we then dynamically select the factors that are most relevant to the current task and most influential for response generation, and incorporate them as natural-language guidance alongside the retrieved history. 

Across four personalization benchmarks and four backbone models, Dynamic PersonaFactor achieves the strongest performance across all reported metrics, consistently outperforming history retrieval and static factor guidance across both open-weight and black-box models. Ablation studies further show that combining task-relevance and generation-influence signals is more effective than using either signal alone or selecting the same number of factors at random, highlighting the importance of request-dependent factor selection. Further analyses show that PersonaFactors share stable low-dimensional structure with user-history effects while providing complementary personalization signals, suggesting that they capture reusable information beyond raw history alone.

Our contributions can be summarized as follows:

\noindent(1) \textbf{Reusable PersonaFactors.}
We introduce a shared set of 13 interpretable personalization factors that capture recurring mechanisms across heterogeneous user interactions and tasks.

\noindent(2) \textbf{Dynamic factor-guided personalization.}
We develop a request-dependent selection method that identifies the factors most relevant to the current task and most influential for generation, and uses them as natural-language guidance.

\noindent(3) \textbf{Empirical validation.}
Experiments across four personalization benchmarks and four backbone models show consistent gains over history retrieval and static factor guidance, while further analyses reveal that PersonaFactors provide complementary signals to user history.

%% file: 2_Related_Work.tex
\section{Related Work}

\noindent\textbf{Personalization Methods.} Existing personalization methods
mainly operate in either model space or natural-language space. Model-space
methods adjust model behavior through fine-tuning, parameter-efficient
adaptation, preference learning, activation steering, or decoding-time control
~\citep{zhang2025proper,choi2025copl,qiu2025latent,zhang2025personalized,
nie2025adaptive,chen2025pad}. They can directly control generation, but
typically require access to model parameters, internal activations, or output
distributions, making them unsuitable for black-box LLMs. Natural-language-space methods instead use user history,
profiles, retrieved records, structured reasoning, or long-term memory as
external context~\citep{su2025personalized,du2026optimizing,kim2026rpm,
noh2026triple,pan2025secom,kang2025memory}. These methods support both
black-box and white-box models, but user preferences often remain entangled in
raw histories or a single profile. PersonaFactors extend this line of work by
providing an interpretable and dynamically composable factor library mined from
large-scale data across personalization scenarios, thereby facilitating more
effective use of user histories.

\noindent\textbf{Personalization Benchmarks and Positioning.} Existing
benchmarks evaluate personalization through task metrics, LLM-as-a-Judge, or
user-specific reward models, covering classification, rating, question
answering, and short- or long-form generation~\citep{salemi2024lamp,
kumar2024longlamp,salemi2025lamp,zollo2024personalllm,su2026copa}. These
benchmarks span diverse tasks, but most studies focus on a single task or
scenario and use user preferences primarily as an evaluation target. CoPA is
closest to our work. It introduces cognitive factors for personalized question
answering, but uses them only as evaluation dimensions ~\citep{su2026copa}. In contrast, PersonaFactors extract
evidence-grounded personalization factors from user histories and responses
across multiple scenarios, then dynamically select and compose relevant
factors for each request, thereby making these factors reusable
representations for generation. Full
descriptions of the related work are provided in
Appendix~\ref{app:related-work-details}.

%% file: 3_Method_yinxia.tex
\section{Method}

\begin{figure}[t]
    \centering
    \includegraphics[width=\textwidth]{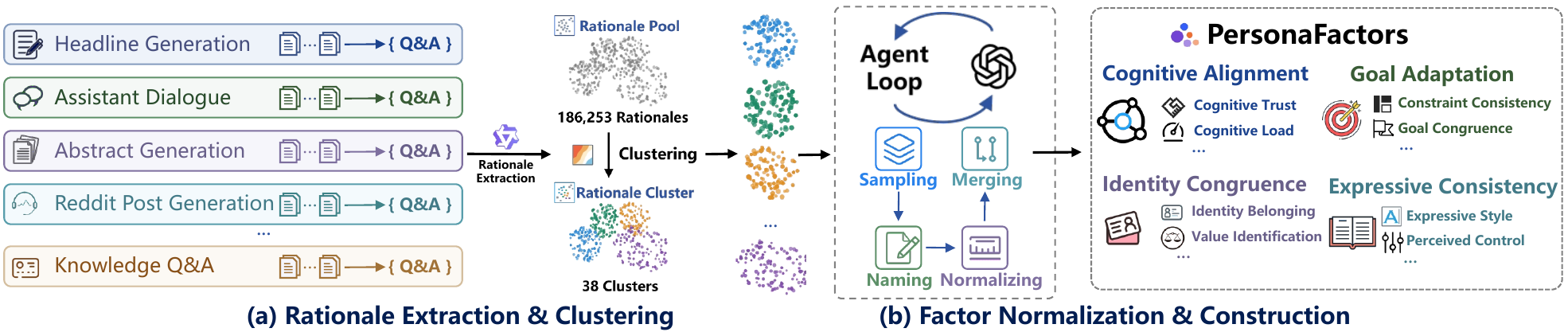}
    \caption{Overview of constructing PersonaFactors. We extract
    personalization rationales from user interactions across heterogeneous
    scenarios, cluster them into candidate factors, and obtain canonical
    PersonaFactors through iterative sampling, naming, merging, and
    normalization.}
    \label{fig:factor-construction}
\end{figure}

\begin{figure}[!htbp]
    \centering
    \includegraphics[width=\textwidth]{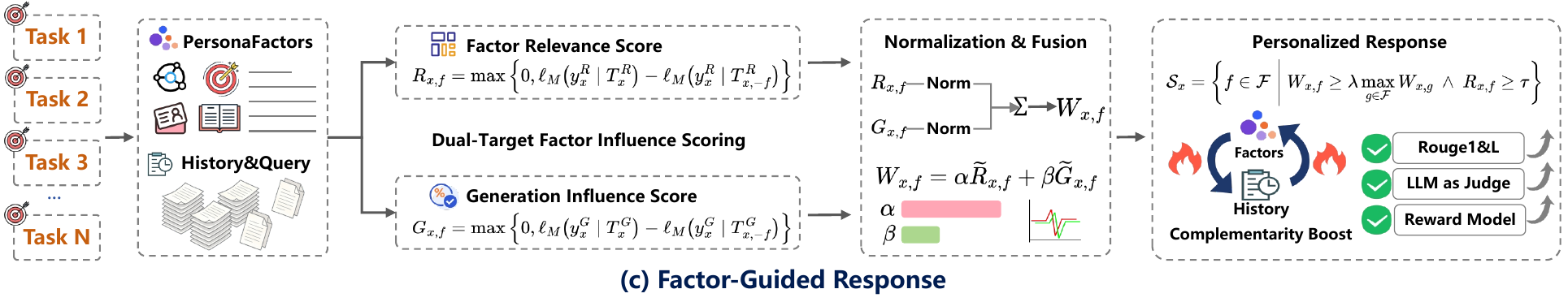}
    \caption{Overview of dynamic factor-guided personalized generation. Given the user
    history and current query, we score each PersonaFactor by its task relevance
    and generation influence, fuse the two signals to select relevant factors,
    and use them to guide personalized response generation.}
    \label{fig:dynamic-factor-personalization}
\end{figure}

We introduce \textbf{PersonaFactors}, reusable personalization mechanisms that
capture how responses should adapt to users across different scenarios. Our
method consists of two stages. We first construct a shared factor library from
heterogeneous user interactions through rationale extraction, clustering, and
mechanism canonicalization (Figure~\ref{fig:factor-construction}). Given this
library, we dynamically select factors relevant to a user's history and current
request, then incorporate them as natural-language guidance for personalized
generation (Figure~\ref{fig:dynamic-factor-personalization}).

\subsection{Cross-Scenario PersonaFactor Construction}

Inspired by the cognitive--affective processing system (CAPS)
theory~\citep{mischel1995cognitive}, we seek recurring personalization
mechanisms whose manifestations vary across situations. We construct a shared
library $\mathcal{F}=\{f_1,\ldots,f_K\}$.
Our construction covers common personalization scenarios, such as
question answering, preference prediction, and short- and long-form text
generation. These scenarios are drawn from the training splits of
LaMP~\citep{salemi2024lamp} and LongLaMP~\citep{kumar2024longlamp}, together
with the full datasets of HiCUPID~\citep{mok2025hicupid} and
CoPA~\citep{su2026copa}.
We randomly sample up to 5,000 instances per constituent task, obtaining
66,018 instances across 14 task-specific subsets.
As illustrated in Figure~\ref{fig:factor-construction}, we first
infer, for each sample, the possible reasons for the user's preference for
the personalized response, grounding these reasons in the user's history and
current query. We then hierarchically cluster these reasons to identify
shared personalization mechanisms across scenarios. Finally, we name and
canonicalize the clusters to distill personalization dimensions and factors
expressed in natural language.
A construction model $M_c$ performs rationale extraction, mechanism naming,
and canonicalization, while a separate model $M_{\mathrm{audit}}$ audits the
resulting factor mapping.

\paragraph{Evidence-grounded Rationale Extraction.}
Each construction instance $x=(h_x,q_x,a_x)$ contains a user history, a
current query, and a personalized response. Rather than  summarizing
the user or the interaction topic, we ask $M_c$ to explain the reasons for
the user's preference for the response:
\begin{equation}
    \mathcal{A}_x=M_c(h_x,q_x,a_x)
    =\{(r_x^j,t_x^j,e_x^j)\}_{j=1}^{k_x}.
\end{equation}

Here, $k_x$ is the number of rationales extracted for instance $x$.
For each rationale, $r_x^j$ describes the reason for the user's preference,
$t_x^j$ identifies its grounding psychological theory, and $e_x^j$ records
supporting input evidence. An interaction may yield multiple rationales,
capturing distinct aspects of personalization. The prompt requires
evidence-grounded explanations and forbids unsupported preferences. Pooling
all rationales gives $\mathcal{A}=\bigcup_x\mathcal{A}_x$, with 186,253
instances.


Two graduate-student annotators independently assess 200 randomly sampled
extraction instances, checking whether the extracted rationales are supported
by the user history, query, and response, use valid theories appropriately,
and provide logically coherent explanations. The evaluation yields a validity
rate of 98\% and inter-annotator agreement of Cohen's $\kappa=0.8455$.
Extraction prompts, examples, and annotation criteria are provided in
Appendices~\ref{app:rationale-extraction-prompt},
\ref{app:rationale-extraction-examples}, and
\ref{app:rationale-manual-inspection}, respectively.

An external rationale extraction evaluation further supports the
effectiveness of our method. We evaluate our rationale extraction module on
ThoughtTrace~\citep{jin2026thoughttrace}, a dataset for inferring the reasons
behind users' choices from given conversation histories. The module recovers
most of these reasons, suggesting that it captures more than surface-level
dialogue actions.
Details are provided in Appendix~\ref{app:reason-recovery}.

\paragraph{Cross-scenario Rationale Clustering.}
We jointly cluster rationales from all construction scenarios to identify
shared adaptation mechanisms. Each rationale is encoded as
$z_x^j=\operatorname{Embed}(t_x^j,r_x^j,e_x^j)$. We apply
UMAP~\citep{mcinnes2018umap} to obtain reduced representations $\tilde z_x^j$,
followed by HDBSCAN~\citep{mcinnes2017hdbscan} to discover candidate clusters.

Let $\mathcal{C}$ denote the candidate cluster set and
$p(c\mid\tilde z_x^j)$ the soft membership probability for cluster $c$,
with $\rho\in(0,1)$ denoting the confidence threshold.
We allow a rationale to support multiple candidate mechanisms through
threshold-based assignment:
\begin{equation}
    \mathcal{C}_x^j
    =\{c\in\mathcal{C}\mid p(c\mid\tilde z_x^j)\geq\rho\}.
\end{equation}
When no cluster meets the primary threshold, we retain a non-noise hard
assignment only if it satisfies a lower membership threshold; otherwise,
the rationale is treated as noise. We select clustering configurations
through a multi-seed search that enforces constraints on coverage, cluster
size, and cross-dataset support, prioritizing structural stability and using
DBCV~\citep{moulavi2014density} as a secondary criterion.
The selected configuration produces 38 candidate clusters. The complete
assignment rules, search procedure, and hyperparameters are provided in
Appendix~\ref{app:rationale-clustering}.

\paragraph{Mechanism Naming and Canonicalization.}
For each cluster, we deduplicate high-membership rationales and select up to
200 representative examples, prioritizing coverage across theoretical labels,
mechanism descriptions, and evidence patterns. The construction model uses
these examples to name the shared adaptation mechanism rather than summarize
surface topics or tasks. We then canonicalize the candidate factors by merging
equivalent or synonymous names for the same concrete mechanism. Each candidate
maps to exactly one canonical factor; uncertain matches remain separate.
An independent audit--repair procedure checks for incorrect or missed merges,
with the full procedure provided in Appendix~\ref{app:factor-naming-prompt}.

This process yields 13 PersonaFactors organized into four dimensions,
summarized in Table~\ref{tab:persona-factor-ontology}. Each factor describes a
reusable user--response alignment property rather than a fixed psychological
trait or a numerical user attribute. For example, \emph{Schema Congruence}
concerns whether an explanation fits the user's knowledge and familiar
concepts, whereas \emph{Constraint Consistency} concerns whether it respects
the user's practical constraints. Complete definitions are provided in
Appendix~\ref{app:persona-factor-definitions}.

\begin{table*}[t]
    \centering
    \small
    \caption{The shared PersonaFactor library.}
    \label{tab:persona-factor-ontology}
    \begin{tabular}{@{}>{\raggedright\arraybackslash}m{0.40\textwidth}
                    >{\raggedright\arraybackslash}m{0.52\textwidth}@{}}
        \hline
        \textbf{Dimension} & \textbf{PersonaFactor} \\
        \hline
        Cognitive and Knowledge Alignment & Cognitive Trust; Schema Congruence;
        Cognitive Consistency; Cognitive Load \\
        Context and Goal Adaptation & Situational Anchoring; Goal Congruence;
        Constraint Consistency \\
        Self and Identity Congruence & Self-Concept Congruence; Identity
        Belonging; Value Identification \\
        \shortstack[l]{Behavioral and Expressive Style\\Consistency} & Behavioral Pattern
        Consistency; Expressive Style Consistency; Perceived Autonomy Control \\
        \hline
    \end{tabular}
\end{table*}

The factor inventory is shared across users and scenarios; the factors used
for a particular request are selected dynamically, as described next.

\subsection{Dynamic Factor-Guided Personalization}
\label{sec:dynamic_factor}

Different requests may require different personalization mechanisms, even for the
same user. Given the shared PersonaFactor library
\(\mathcal{F}=\{f_1,\ldots,f_K\}\), we  select a task-specific subset
for each inference instance \(x=(h_x,q_x)\), where \(h_x\) denotes the retrieved
user history and \(q_x\) the current request.

Our key idea is to estimate the importance of each factor in \(\mathcal{F}\) through a \emph{leave-one-out} test. We first run the backbone model \(M\), which is also used for personalized response generation, with the complete factor library \(\mathcal{F}\) to obtain fixed reference outputs. We then remove each factor individually and measure how much the model's support for the same outputs decreases. A larger decrease indicates that the removed factor is more influential for the current request. The selected factors are then used together with the retrieved user history to guide the final response generation.

\paragraph{Reference Targets.}
For each instance, we obtain two reference outputs using the complete factor library. The two prompts contain the same factor library, user history, and current request, and differ only in their instructions. The \emph{relevance prompt} (denoted by $R$) asks the model to identify which personalization factors are relevant to the current request, while the \emph{generation prompt} (denoted by $G$) asks the model to produce a personalized response.

Formally, for target type $t\in\{R,G\}$, we construct $T_x^t = [\mathcal{F}; I^t; h_x; q_x]$, where $I^R$ is the relevance instruction and $I^G$ is the response-generation instruction. The backbone model produces the relevance judgment $y_x^R$ and a draft personalized response $y_x^G$:
\begin{equation}
    y_x^R = M(T_x^R),
    \qquad
    y_x^G = M(T_x^G),
\end{equation}
where both outputs are generated once and  kept fixed during the subsequent leave-one-out scoring.

\paragraph{Leave-One-Factor-Out Influence.}
For each factor \(f\in\mathcal{F}\), we remove only \(f\) from the prompt while
keeping the instruction, history, query, and reference target unchanged.
Denote $T_{x,\backslash f}^t=[\mathcal{F}\setminus\{f\}; I^t; h_x; q_x]$. 
We define the influence of \(f\) as the decrease in log-likelihood assigned to
the same fixed target after removing that factor:
\begin{equation}
    \Delta_{x,f}^t
    =
    \left[
    \log p_M(y_x^t\mid T_x^t)
    -
    \log p_M(y_x^t\mid T_{x,\backslash f}^t)
    \right]_+,
    \qquad t\in\{R,G\},
    \label{eq:factor_influence}
\end{equation}
where \([u]_+=\max(0,u)\).

This comparison isolates the contribution of a single factor: if removing
\(f\) substantially lowers the likelihood of the fixed target, then \(f\)
provides strong support for that target. If removing \(f\) leaves the target
likelihood unchanged or increases it, its positive influence is set to zero.
We denote
\[
    R_{x,f}=\Delta_{x,f}^R,
    \qquad
    G_{x,f}=\Delta_{x,f}^G,
\]
where \(R_{x,f}\) measures influence on the relevance judgment and \(G_{x,f}\)
measures influence on the draft response. 

\paragraph{Influence Fusion and Factor Selection.}
Because the two influence signals have different scales, we normalize them independently across factors within each instance, obtaining \(\widetilde{R}_{x,f}\) and \(\widetilde{G}_{x,f}\). We then combine them as
\begin{equation}
    W_{x,f}
    =
    \alpha \widetilde{R}_{x,f}
    +
    \beta \widetilde{G}_{x,f},
    \qquad
    \alpha+\beta=1.
    \label{eq:factor_fusion}
\end{equation}


We use a two-criterion selection rule. A factor is retained if its fused influence is within a fraction \(\lambda\) of the strongest factor for the current instance and its relevance influence exceeds an absolute threshold \(\tau\).
\begin{equation}
    S_x
    =
    \left\{
    f\in\mathcal{F}
    \;\middle|\;
    W_{x,f}\ge
    \lambda \max_{g\in\mathcal{F}} W_{x,g},
    \;
    R_{x,f}\ge\tau
    \right\}.
    \label{eq:factor_selection}
\end{equation}
The relative criterion allows the number of selected factors to adapt to each request, while the relevance threshold avoids retaining factors when all influence scores are weak.

\paragraph{Factor-Guided Response Generation.}
After selection, only the factors in \(S_x\) are included as natural-language guidance together with the retrieved history and current query. The model then generates the final personalized response:
\[
    \hat{y}_x
    \sim
    M([S_x; I^{\mathrm{gen}}; h_x; q_x]).
\]
The factors specify which dimensions of personalization should be emphasized, while the retrieved history provides the user-specific evidence needed to instantiate them. The influence scores are used only for factor selection and are not passed to the model as continuous generation weights.

\paragraph{Prompt-Based Selection for Black-Box Models.}
The score-based procedure above requires access to sequence likelihoods. When such scores are unavailable, as in black-box APIs, we retain the same PersonaFactor representation and downstream generation interface but replace
leave-one-factor-out scoring with a prompt-based selector. Given
\((h_x,q_x,\mathcal{F})\), the selector directly predicts a task-specific subset
\(S_x\), which is then used by the same factor-guided response generation stage.
Thus, the two realizations differ only in how the factor subset is inferred.
The prompt-based selector does not use Equation~\ref{eq:factor_influence} or
Equation~\ref{eq:factor_selection}. The corresponding prompt is provided in
Appendix~\ref{app:dynamic-factor-selection-prompt}.

%% file: 4_Experiments.tex
\section{Experiments}

\subsection{Experimental Setup}

\paragraph{Models.}
We evaluate PersonaFactor-guided personalization on four backbones:
the API-based DeepSeek-V4-Pro~\citep{deepseekai2026v4} and
Gemini-2.5-Flash~\citep{comanici2025gemini}, and the open-weight
Qwen3-8B-Instruct~\citep{yang2025qwen3} and
Gemma-3-4B~\citep{team2025gemma}. We use the score-based selector when
sequence likelihoods are available and the prompt-based selector for black-box
APIs without likelihood access. Implementation details are provided
in Appendix~\ref{app:implementation-details}.

\paragraph{Benchmarks.}
\noindent We evaluate on four personalized generation benchmarks: LaMP~\citep{salemi2024lamp}, LongLaMP~\citep{kumar2024longlamp}, LaMP-QA~\citep{salemi2025lamp}, and PersonalLLM~\citep{zollo2024personalllm}. LaMP and LongLaMP are used to construct PersonaFactors, with their validation splits serving for evaluation, whereas LaMP-QA and PersonalLLM are reserved for downstream testing. LaMP covers news headline generation, scholarly title generation, and tweet paraphrasing, while LongLaMP covers abstract, product review, and topic generation; both use user histories and current inputs and are evaluated with ROUGE-1 and ROUGE-L. LaMP-QA assesses personalized long-form question answering across Art and Entertainment, Lifestyle and Personal Development, and Society and Culture using an LLM-as-a-Judge protocol. PersonalLLM measures open-domain preference alignment with heterogeneous reward models. Together, these benchmarks span diverse tasks, domains, and evaluation paradigms, enabling assessment of cross-task generalization and robustness. Detailed statistics are provided in Appendix~\ref{app:benchmark-statistics}.

\paragraph{Baselines.}
We compare six methods that differ in how they use user history and personalization factors. \textbf{Direct} generates a response from the current input without access to user history. \textbf{RAG} retrieves history items relevant to the current input and provides them as context to the model. \textbf{RAG-SelfFactor} uses the current query and retrieved user history to generate query-specific factors that augment the retrieved context. The corresponding prompt is provided in Appendix~\ref{app:rag-selffactor-prompt}. \textbf{RAG-CoPA} augments retrieval-based personalization with CoPA factors~\citep{su2026copa}. \textbf{Static PersonaFactor} builds on the same RAG procedure and uses a fixed set of pre-constructed personalization factors. \textbf{Dynamic PersonaFactor} also builds on RAG, but selects the factors relevant to each current task dynamically.

\subsection{Main Results}

As shown in Table~\ref{tab:main-results}, Dynamic PersonaFactor
achieves the best performance on every reported metric across all four
backbone models and four benchmarks, demonstrating the
applicability of our method across diverse personalization scenarios. Detailed results are provided in
Appendix~\ref{app:detailed-main-results}.

\begin{table}[!t]
    \centering
    \small
    \fontfamily{ptm}\selectfont
    \setlength{\tabcolsep}{2.5pt}
    \renewcommand{\arraystretch}{1.08}
    \caption{Main results across models and personalization methods. LaMP
    and LongLaMP report macro-averaged ROUGE-1/ROUGE-L over their evaluated
    tasks; LaMP-QA uses an LLM-as-a-Judge score, and PersonalLLM uses a reward
    model score. Best results within each model are boldfaced. Results are
    reported as averages over three independent runs.}
    \label{tab:main-results}
    \begin{tabular}{@{}llcccccc@{}}
        \toprule
        \multirow{2}{*}{\textbf{Model}} & \multirow{2}{*}{\textbf{Method}} & \multicolumn{2}{c}{\textbf{LaMP}} & \multicolumn{2}{c}{\textbf{LongLaMP}} & \textbf{LaMP-QA} & \textbf{PersonalLLM} \\
        \cmidrule(lr){3-4} \cmidrule(lr){5-6} \cmidrule(lr){7-7} \cmidrule(lr){8-8}
        & & R-1 & R-L & R-1 & R-L & {\scriptsize LLM Judge} & {\scriptsize Reward Model} \\
        \midrule
        \multirow{6}{*}{DeepSeek-V4-Pro}
        & Direct & 0.2932 & 0.2580 & 0.2125 & 0.1242 & 0.4915 & -0.9013 \\
        & RAG & 0.3400 & 0.3004 & 0.3003 & 0.1538 & 0.6503 & 0.4049 \\
        & RAG-SelfFactor & 0.3325 & 0.2904 & 0.2891 & 0.1561 & 0.5940 & -0.3861 \\
        & RAG-CoPA & 0.3456 & 0.3062 & 0.3082 & 0.1565 & 0.7196 & 0.4642 \\
        & Static PersonaFactor & 0.3495 & 0.3076 & 0.3141 & 0.1581 & 0.7418 & 0.4795 \\
        & Dynamic PersonaFactor & \textbf{0.3560} & \textbf{0.3140} & \textbf{0.3645} & \textbf{0.1935} & \textbf{0.7575} & \textbf{0.4915} \\
        \midrule
        \multirow{6}{*}{Gemini-2.5-Flash}
        & Direct & 0.3286 & 0.2918 & 0.2305 & 0.1309 & 0.4354 & -0.4993 \\
        & RAG & 0.4010 & 0.3614 & 0.3121 & 0.1615 & 0.5748 & 0.2426 \\
        & RAG-SelfFactor & 0.3809 & 0.3419 & 0.2689 & 0.1394 & 0.5178 & -0.6971 \\
        & RAG-CoPA & 0.4027 & 0.3637 & 0.3395 & 0.1746 & 0.5810 & 0.2689 \\
        & Static PersonaFactor & 0.4100 & 0.3696 & 0.3461 & 0.1789 & 0.5914 & 0.2822 \\
        & Dynamic PersonaFactor & \textbf{0.4225} & \textbf{0.3761} & \textbf{0.3888} & \textbf{0.2051} & \textbf{0.6141} & \textbf{0.3022} \\
        \midrule
        \multirow{6}{*}{Qwen3-8B}
        & Direct & 0.3145 & 0.2785 & 0.2151 & 0.1287 & 0.3271 & -1.2800 \\
        & RAG & 0.3249 & 0.2877 & 0.2581 & 0.1437 & 0.4159 & -0.2778 \\
        & RAG-SelfFactor & 0.3065 & 0.2674 & 0.3017 & 0.1594 & 0.3684 & -0.3638 \\
        & RAG-CoPA & 0.3273 & 0.2899 & 0.2657 & 0.1457 & 0.4303 & -0.1901 \\
        & Static PersonaFactor & 0.3276 & 0.2892 & 0.2719 & 0.1476 & 0.4328 & -0.1876 \\
        & Dynamic PersonaFactor & \textbf{0.3336} & \textbf{0.2956} & \textbf{0.3377} & \textbf{0.1859} & \textbf{0.4417} & \textbf{-0.1722} \\
        \midrule
        \multirow{6}{*}{Gemma-3-4B}
        & Direct & 0.2522 & 0.2211 & 0.2005 & 0.1194 & 0.3762 & -0.2998 \\
        & RAG & 0.2706 & 0.2364 & 0.2319 & 0.1328 & 0.4450 & -0.2414 \\
        & RAG-SelfFactor & 0.2645 & 0.2333 & 0.2029 & 0.1156 & 0.3995 & -0.2562 \\
        & RAG-CoPA & 0.2695 & 0.2354 & 0.2280 & 0.1312 & 0.4981 & -0.2012 \\
        & Static PersonaFactor & 0.2710 & 0.2353 & 0.2346 & 0.1339 & 0.4944 & -0.1977 \\
        & Dynamic PersonaFactor & \textbf{0.2818} & \textbf{0.2490} & \textbf{0.2943} & \textbf{0.1651} & \textbf{0.5107} & \textbf{-0.1836} \\
        \bottomrule
    \end{tabular}
\end{table}

Compared with Static PersonaFactor, Dynamic PersonaFactor
consistently improves performance across all four benchmarks, highlighting the importance of
task-adaptive factor selection. Different inputs may require different subsets
of a user's preferences, whereas injecting the entire factor inventory can
introduce redundant or mutually competing constraints. Dynamic selection
mitigates this issue by concentrating personalization signals on the factors
most relevant to the current task.

Even without dynamic selection, Static PersonaFactor generally
outperforms RAG-CoPA across backbone models and evaluation protocols. This
advantage supports the effectiveness of our hierarchical, cross-scenario
factor extraction in constructing a reusable personalization factor library.
RAG-SelfFactor, by contrast, generally underperforms methods using
pre-constructed factor libraries and sometimes even performs worse than RAG.
These results highlight the limitations of asking an LLM to generate
personalization factors on the fly during inference and underscore the need
for constructing a reusable factor library.

The overall advantage of Dynamic PersonaFactor holds for both
black-box API models and open-weight models, despite their different inference
interfaces and parameter scales. This
cross-model consistency suggests that the benefit of factorized, dynamically
selected personalization is not tied to a particular backbone or access mode,
providing evidence that the personalization mechanism transfers across model
families and access settings.
\subsection{Dynamic Factor Selector Ablation}
\label{sec:dynamic-selector-ablation}

To quantify the contribution of the signals used in dynamic factor
construction, we perform a controlled ablation on Qwen3-8B with LaMP-QA. We
compare the full strategy, which combines task-relevance influence $R_{x,f}$ and
generation influence $G_{x,f}$, with $R_{x,f}$-only and $G_{x,f}$-only variants. Random-$k$
selects the same number of factors as the full strategy. All conditions
share the same user history, generation prompt, decoding configuration, and
factor budget, allowing Random-$k$ to control for gains attributable solely to
the number of injected factor descriptions.

Table~\ref{tab:dynamic-selector-ablation} reports results across the three
LaMP-QA domains. Dynamic PersonaFactor achieves the highest score in every
domain. The $R_{x,f}$-only variant is the strongest single-signal strategy, while
both single-signal variants and Random-$k$ consistently trail the complete
strategy. Its margin over Random-$k$ indicates that the gains arise from
selecting relevant factors rather than from increasing factor count alone.
The gap between the two single-signal variants further suggests that
task-relevance influence is more informative than generation influence in this
setting. Results for Gemma-3-4B are provided in
Appendix~\ref{app:gemma-dynamic-selector-ablation}. We also compare the
prompt-based selector realizations on both open-weight models,
the results are reported in Appendix~\ref{app:open-model-two-stage}.

\begin{table}[!htbp]
    \centering
    \small
    \setlength{\tabcolsep}{4.5pt}
    \renewcommand{\arraystretch}{1.12}
    \caption{Ablation of the dynamic factor selector on Qwen3-8B using
    LaMP-QA. The three columns report the benchmark's LLM-as-a-Judge score
    for each domain. Higher is better.}
    \label{tab:dynamic-selector-ablation}
    \begin{tabular}{@{}lccc@{}}
        \toprule
        \textbf{Selector}
            & \shortstack{Art \&\\Entertainment}
            & \shortstack{Lifestyle \&\\Personal Development}
            & \shortstack{Society \&\\Culture} \\
        \midrule
        Dynamic PersonaFactor & \textbf{0.3570} & \textbf{0.4542}
            & \textbf{0.5139} \\
        $R_{x,f}$-only & 0.3457 & 0.4502 & 0.5099 \\
        $G_{x,f}$-only & 0.3312 & 0.4427 & 0.5023 \\
        Random-$k$           & 0.3396 & 0.4398 & 0.5056 \\
        \bottomrule
    \end{tabular}
\end{table}

\subsection{Effect of the Number of History Items}
\label{sec:history-count-effect}

\begin{figure}[!htbp]
    \centering
    \includegraphics[width=\linewidth]{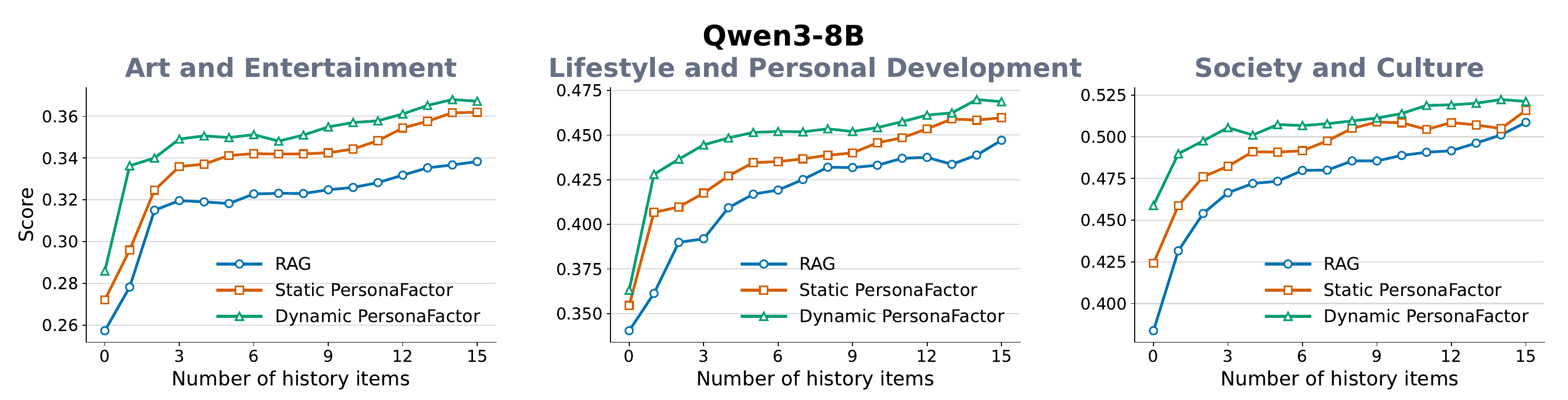}
    \caption{LaMP-QA performance as a function of the number of user-history
    items for Qwen3-8B.}
    \label{fig:history-count-effect}
\end{figure}

We vary the number of user-history items and compare RAG, Static PersonaFactor,
and Dynamic PersonaFactor on Qwen3-8B to assess sensitivity to history size,
especially for sparse user profiles. As shown in
Figure~\ref{fig:history-count-effect}, performance improves rapidly with the
first few items and then gradually saturates. Dynamic PersonaFactor outperforms
both baselines across nearly all history lengths, with the clearest advantage
in the low-history regime, suggesting its effectiveness in cold-start settings.

At larger history sizes, mild non-monotonic fluctuations show that additional
interactions do not guarantee continued gains, as irrelevant or competing
cross-scenario preferences may dilute task-relevant signals. Overall, the
results suggest that Dynamic PersonaFactor benefits from selectively structuring
historical evidence rather than merely increasing context length. Corresponding
Gemma-3-4B results are provided in
Appendix~\ref{app:gemma-history-count-effect}.

\subsection{History and PersonaFactor Effects in Representation Space}
\label{sec:low-dimensional-analysis}
\label{sec:combined-effects-analysis}

We further investigate how PersonaFactors influence the backbone model's internal
representations. Specifically, we address two questions: (1) whether PersonaFactors
and user history induce representation shifts along similar directions, and (2) whether
PersonaFactors provide additional information when combined with user history. We compare
\emph{History}, \emph{PersonaFactor}, and \emph{History+PersonaFactor} on Qwen3-8B
across LaMP and LongLaMP, using Static PersonaFactor to avoid variation from dynamic
factor selection. For each sample and transformer layer, we extract the final
prompt-token hidden state and compute the corresponding representation shifts;
formal definitions and additional methodological details are provided in
Appendix~\ref{app:low-dimensional-analysis}.

\paragraph{Do History and PersonaFactors induce similar representation shifts?}

\begin{figure}[!htbp]
    \centering
    \includegraphics[width=\linewidth]{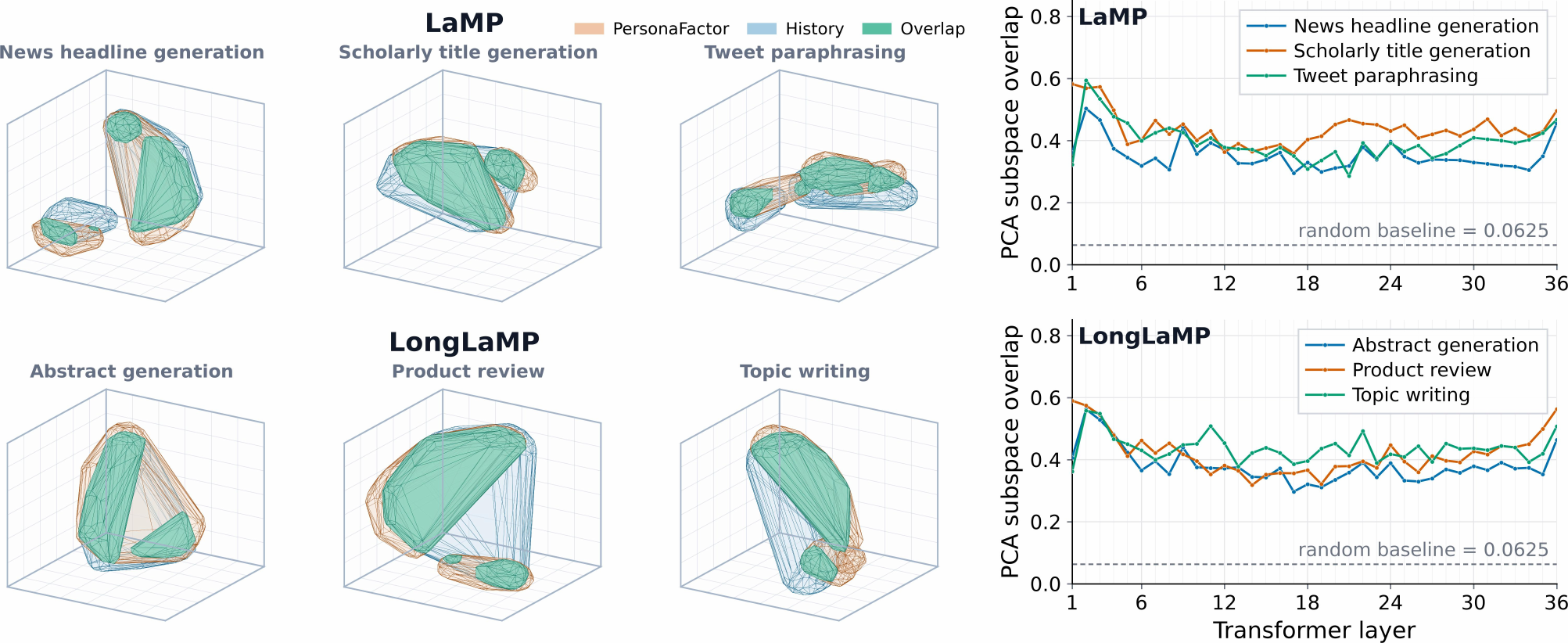}
    \caption{Low-dimensional overlap between History and PersonaFactor effects
    across LaMP and LongLaMP tasks and layers for Qwen3-8B.}
    \label{fig:low-dimensional-overlap}
\end{figure}

To test whether History and PersonaFactor induce shifts along similar directions, we
project their normalized difference vectors into a shared 128-dimensional space and
measure the overlap between their top-3 subspaces. As shown in
Figure~\ref{fig:low-dimensional-overlap}, the mean task-level overlap on Qwen3-8B
is $0.236$, substantially above the random-subspace expectation of $0.0234$.
The layer-wise results show the same trend. The Gemma-3-4B results show the same
trend as well. Details are provided in
Appendix~\ref{app:low-dimensional-analysis}. These results show that History and
PersonaFactor induce shifts along partially shared low-dimensional directions.

\paragraph{Do PersonaFactors provide information beyond History?}

\begin{figure}[!htbp]
    \centering
    \includegraphics[width=\linewidth]{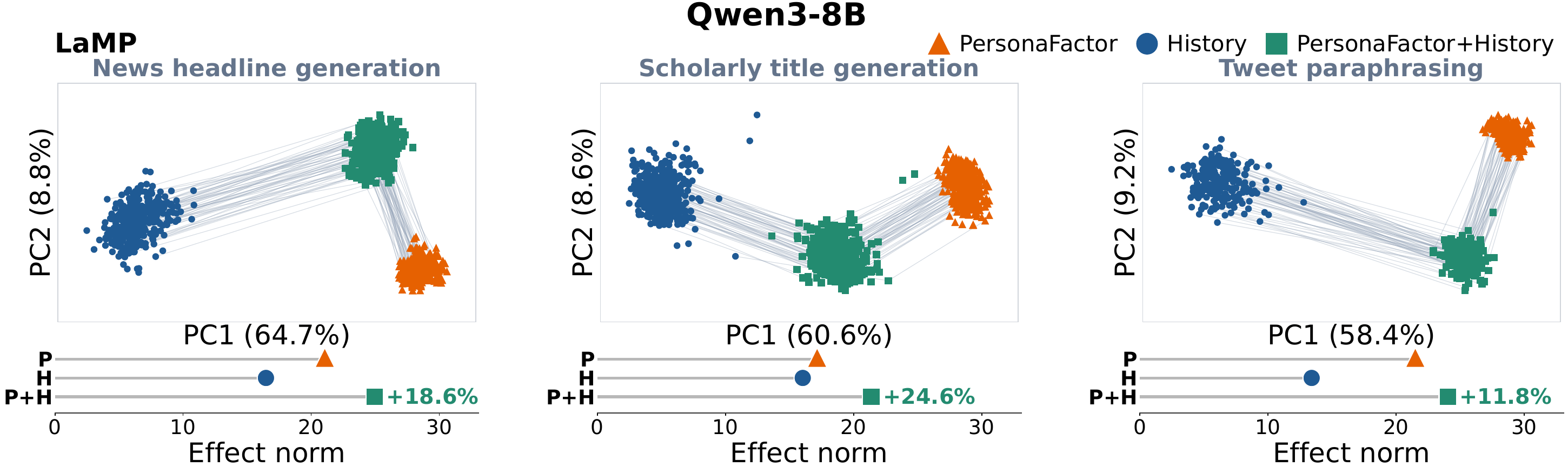}
    \caption{Representation shifts induced by History, PersonaFactor, and their
    joint condition for Qwen3-8B on LaMP. Top: normalized shifts in a shared
    two-dimensional PCA space. Lines connect matched observations across input
    conditions. Bottom: mean unnormalized $L_2$ norms with bootstrap 95\%
    confidence intervals.}
    \label{fig:combined-effects-qwen-lamp}
\end{figure}

To assess whether PersonaFactors provide information beyond History, we compare the
three conditions in Figure~\ref{fig:combined-effects-qwen-lamp}. In the shared
two-dimensional PCA space, the joint condition occupies a distinct region from
either individual condition. Its unnormalized shifts also have the largest mean
$L_2$ norm across all three LaMP tasks. The corresponding LongLaMP results are
provided in Appendix~\ref{app:combined-effects-analysis}, where the same trend is
observed. Thus, History and PersonaFactor share part of their representation
structure, while their combination produces additional changes.

%% file: 5_Conclusion.tex
\section{Conclusion}

We introduced PersonaFactors, a framework that distills reusable and
interpretable personalization factors from heterogeneous user interactions and
selects them dynamically for each request. Across four benchmarks and four
backbone models, Dynamic PersonaFactor achieves the best score on every
reported metric, consistently outperforming history retrieval and static factor
guidance. Selector ablations show that combining task-relevance and
generation-influence signals is more effective than either signal alone or
matched-size random selection. Representation analyses using Static
PersonaFactor show partially shared low-dimensional structure between factor-
and history-induced shifts, while their combination produces additional
representation changes. Together, these results suggest that organizing user history through reusable,
request-specific factors provides a practical way to improve personalization
across tasks and model families. Limitations are discussed in
Appendix~\ref{app:limitations}.

%% file: 11_AI_Use_Statement.tex
\section*{AI Use Statement}

Generative AI tools were used to assist with language editing, translation,
and formatting during manuscript preparation. The authors reviewed and
verified all AI-assisted text and remain responsible for the final manuscript,
including its claims, analyses, and artifacts. The experimental methodology,
implementation, data processing, and reported results were independently
checked by the authors.

%% file: 12_Reproducibility_Statement.tex
\section*{Reproducibility Statement}

We provide the formal method definitions in Section~3 and the complete
experimental protocol, model settings, and evaluation details in
Section~4 and Appendix~\ref{app:implementation-details}. The construction
prompts, factor definitions, clustering procedure, selection parameters, and
detailed results are documented in the appendix. All reported scores are
averaged over three independent runs. Code and supplementary materials will be
released in an anonymized form for review and made public upon publication.

%% file: 7_Ethics_Statement.tex
\section*{Ethics Statement}

This work uses only publicly available benchmark datasets and does not collect
or process any new user data or personally identifiable information. It
therefore involves no new human-subjects research. The model-generated factors
are intended solely as textual guidance for personalized generation and should
not be interpreted as validated or reliable assessments of an individual's
psychological traits, identity, or preferences.

%% file: 9_Appendix.tex
\setcounter{figure}{0}
\renewcommand{\thefigure}{A\arabic{figure}}
\renewcommand{\theHfigure}{A\arabic{figure}}
\setcounter{table}{0}
\renewcommand{\thetable}{A.\arabic{table}}
\renewcommand{\theHtable}{A.\arabic{table}}
\setcounter{algorithm}{0}
\renewcommand{\thealgorithm}{A\arabic{algorithm}}
\renewcommand{\theHalgorithm}{A\arabic{algorithm}}

\section{Related Work Details}
\label{app:related-work-details}

\noindent\textbf{Model-Space Personalization.} Model-space personalization encodes user information in model parameters or
internal representations. Prior work adapts model behavior to individual users
through user-level fine-tuning, parameter-efficient adaptation, collaborative
preference learning, or inference-time control. PROPER and CoPL respectively
use progressive adaptation and collaborative preference learning for user
modeling~\citep{zhang2025proper,choi2025copl}, while Qiu et al. model
personalized differences between users in a latent space~\citep{qiu2025latent}.
Other approaches steer model behavior through activation directions or adjust
preference-aware output distributions at decoding time~\citep{zhang2025personalized,nie2025adaptive,chen2025pad}.
These methods directly control generation, but typically require access to
model parameters, activations, or output probabilities.

\noindent\textbf{Natural-Language-Space Personalization.} Natural-language-space personalization instead injects user history as external
context through in-context learning. Retrieval-augmented approaches select
behavioral records relevant to the current request, while profile-based methods
summarize dispersed interactions as explicit natural-language preferences.
Long-term memory systems organize and retrieve user information for
personalized agents~\citep{pan2025secom,kang2025memory}. Su et al. generate
and compress user profiles for personalized question answering~\citep{su2025personalized},
whereas Du et al. optimize the profile used for generation with contextual
bandits~\citep{du2026optimizing}. RPM and TRIPLE further use structured
reasoning or behavioral profiles to guide personalized generation~\citep{kim2026rpm,noh2026triple}.
These methods preserve user information in
a readable form and do not require modifying the base model.

\noindent\textbf{Personalization Benchmarks.} Personalization benchmarks employ evaluation signals at different levels to
measure whether a model adapts to a user. LaMP covers personalized
classification, rating, and text generation, using conventional classification,
regression, and reference-based generation metrics~\citep{salemi2024lamp}.
LongLaMP extends this setting to personalized long-form text
generation~\citep{kumar2024longlamp}. Such evaluations enable standardized
comparisons but characterize user preferences only indirectly. LaMP-QA targets
personalized long-form question answering without reference answers and uses an
LLM-as-a-Judge to assess whether a response addresses user-relevant aspects
derived from the question~\citep{salemi2025lamp}. PersonalLLM ensembles reward
models to simulate diverse user preferences and evaluates personalized
generation using the resulting user-specific rewards~\citep{zollo2024personalllm}.
CoPA develops a cognitive-factor-based evaluation framework, using an
LLM-as-a-Judge to score responses along each factor and assess their alignment
with users' cognitive preferences~\citep{su2026copa}.

Among these works, CoPA is most closely related to ours. It distills
personalization factors from a cognitive perspective, but uses them only as
evaluation dimensions in a cognitive question-answering setting. It neither
examines whether factors are shared and transferable across scenarios nor uses
them to guide generation. In contrast, our method extracts reusable
personalization factors from multi-scenario user histories and dynamically
selects and composes relevant factors for each request. The resulting
natural-language guidance enables personalized generation across scenarios.

\section{PersonaFactor Method Details}

\subsection{Rationale Extraction Prompt}
\label{app:rationale-extraction-prompt}

Figure~\ref{fig:rationale-extraction-prompt} shows the prompt used for
rationale extraction.

\begin{figure}[!t]
    \small
    \centering
    \begin{tcolorbox}[colframe=RoyalBlue, colback=white]
    \textbf{System:}
    \\
    You are an expert in personalized response acceptance analysis.
    \\
    \vspace{1mm}

    You must explain why a user would accept a specific personalized answer
    based only on the provided user history, current question, answer, and
    optional comparison context.
    \\
    \vspace{1mm}

    Output must be strict JSON in this format:
    \\
    \{\\
      \hspace*{1em}"reasons": [\\
        \hspace*{2em}\{\\
          \hspace*{3em}"reason": "",\\
          \hspace*{3em}"theory": "",\\
          \hspace*{3em}"explanation": ""\\
        \hspace*{2em}\},\\
        \hspace*{2em}\{\\
          \hspace*{3em}"reason": "",\\
          \hspace*{3em}"theory": "",\\
          \hspace*{3em}"explanation": ""\\
        \hspace*{2em}\}\\
      \hspace*{1em}],\\
      \hspace*{1em}"global\_explanation": ""\\
    \}
    \\
    \vspace{1mm}

    Rules:
    \\
    1. Return one or more reasons.
    \\
    2. Generate multiple reasons when the evidence supports multiple distinct
    explanations.
    \\
    3. Each ``reason'' must be concise and no more than five words.
    \\
    4. Each ``theory'' must use an educational or psychological framework term.
    \\
    5. Each ``explanation'' must be grounded in the provided fields.
    \\
    6. Do not invent user preferences or missing evidence.
    \\
    7. Return JSON only.
    \\
    \vspace{1mm}

    \textbf{User:}
    \\
    User history:
    \\
    \{user\_history\}
    \\
    \vspace{1mm}

    Current question:
    \\
    \{current\_question\}
    \\
    \vspace{1mm}

    Personalized answer:
    \\
    \{personalized\_answer\}
    \end{tcolorbox}
    \caption{Rationale Extraction Prompt.}
    \label{fig:rationale-extraction-prompt}
\end{figure}

\subsection{Rationale Extraction Examples}
\label{app:rationale-extraction-examples}

Tables~\ref{tab:rationale-example-longlamp} and
\ref{tab:rationale-example-lamp7} illustrate how the extraction procedure maps
user history, a current task, and a personalized answer to structured
rationales. Long inputs and answers are abridged with $[\ldots]$.

\begin{table}[!htbp]
    \centering
    \footnotesize
    \setlength{\tabcolsep}{4pt}
    \renewcommand{\arraystretch}{1.12}
    \begin{tabular}{>{\raggedright\arraybackslash}m{0.09\textwidth} >{\raggedright\arraybackslash}p{0.17\textwidth} >{\raggedright\arraybackslash}p{0.66\textwidth}}
        \toprule
        \textbf{Type} & \textbf{Component} & \textbf{Content} \\
        \midrule
        \textbf{Input} & \multicolumn{2}{l}{%
        \begin{tabular}{@{}>{\raggedright\arraybackslash}m{0.17\textwidth} >{\raggedright\arraybackslash}m{0.66\textwidth}@{}}
        Current task & Write a personalized review of a Hamilton Beach two-slice
        toaster, given a 5.0 rating and the summary ``A great general-purpose
        toaster with extra perqs for bagel lovers'' $[\ldots]$. \\
        User history & Prior reviews describe upgrading from inexpensive cookware
        to professional kitchen equipment and emphasize even heating, durability,
        and appearance: ``I never realized how much of a difference REAL kitchen
        gear makes $[\ldots]$'' Other reviews value precise measurements, strong
        motor performance, absorbency, softness, and product design $[\ldots]$. \\
        Answer & ``To be candid, I haven't owned or regularly used a
        toaster in about twenty years $[\ldots]$ It's a sturdy design with a
        nice-looking, soft-industrial aesthetic $[\ldots]$ Once I know what
        settings work for the things I toast, I never have to guess again
        $[\ldots]$ the `bagel' specialty function $[\ldots]$ is quite welcome, as
        is the `defrost' function $[\ldots]$ Auto-shutoff prevents burning in case
        of jams $[\ldots]$.'' \\
        \end{tabular}} \\
        \midrule
        \textbf{Output} & \multicolumn{2}{l}{%
        \begin{tabular}{@{}>{\raggedright\arraybackslash}m{0.17\textwidth} >{\raggedright\arraybackslash}m{0.66\textwidth}@{}}
        Rationale 1 & \textbf{Reason:} Matches user's upgrade mindset.
        \textbf{Theory:} Cognitive Consistency Theory.
        \textbf{Explanation:} The user consistently reviews products where they
        acknowledge upgrading from basic/budget items to quality gear (cookware,
        towels, mixer). The answer mirrors this pattern by admitting past lack of
        toaster use and praising the upgrade to modern features. \\
        Rationale 2 & \textbf{Reason:} Aligns with practical value focus.
        \textbf{Theory:} Utilitarian Value Perception.
        \textbf{Explanation:} User history shows appreciation for functional
        improvements (even cooking, precise measurements, absorbent towels). The
        answer highlights practical benefits like auto-shutoff, defrost function,
        and consistent toasting results. \\
        Rationale 3 & \textbf{Reason:} Reflects aesthetic appreciation.
        \textbf{Theory:} Aesthetic-Usability Effect.
        \textbf{Explanation:} User frequently mentions product appearance
        (handsome cookware, attractive mixer, good-looking towels). The answer
        notes the `soft-industrial aesthetic' and how it looks great in the
        kitchen. \\
        \end{tabular}} \\
        \bottomrule
    \end{tabular}
    \caption{Rationale extraction example from LongLaMP product-review generation.}
    \label{tab:rationale-example-longlamp}
\end{table}

\begin{table}[!htbp]
    \centering
    \footnotesize
    \setlength{\tabcolsep}{4pt}
    \renewcommand{\arraystretch}{1.12}
    \begin{tabular}{>{\raggedright\arraybackslash}m{0.09\textwidth} >{\raggedright\arraybackslash}p{0.17\textwidth} >{\raggedright\arraybackslash}p{0.66\textwidth}}
        \toprule
        \textbf{Type} & \textbf{Component} & \textbf{Content} \\
        \midrule
        \textbf{Input} & \multicolumn{2}{l}{%
        \begin{tabular}{@{}>{\raggedright\arraybackslash}m{0.17\textwidth} >{\raggedright\arraybackslash}m{0.66\textwidth}@{}}
        Current task & ``Paraphrase the following tweet without any explanation
        before or after it: @morphiine, it's amusing that you label ohfukk.com as
        a new venture when it's simply for entertainment.'' \\
        User history & ``@purelynarcotic Ohh.. \#twpp is falling silent.. I was
        waiting to track ur US version.. Next time maybe.. Btw sunny california
        sounds good''; ``@morphiine kya karen yaar.. so many clothes here and
        there.. $[\ldots]$''; ``Ohh the gprs is not so constant on this highway..''
        $[\ldots]$. \\
        Answer & ``@morphiine U call ohfukk.com as a new venture :p
        Thts funny.. Its just fun'' \\
        \end{tabular}} \\
        \midrule
        \textbf{Output} & \multicolumn{2}{l}{%
        \begin{tabular}{@{}>{\raggedright\arraybackslash}m{0.17\textwidth} >{\raggedright\arraybackslash}m{0.66\textwidth}@{}}
        Rationale 1 & \textbf{Reason:} Matches user's casual tone.
        \textbf{Theory:} Tone Congruence Theory.
        \textbf{Explanation:} The user's history shows informal, conversational
        language with emojis and abbreviations (e.g., `Btw', `kya karen yaar').
        The answer mirrors this style with `:p' and `Thats funny', aligning with
        the user's communication preferences. \\
        Rationale 2 & \textbf{Reason:} Reflects user's humor style.
        \textbf{Theory:} Humor Alignment Principle.
        \textbf{Explanation:} The user frequently uses lighthearted expressions
        (e.g., `Ohh..', `Ohh the gprs'). The answer's playful tone (`U call',
        `just fun') resonates with the user's established humor pattern. \\
        \end{tabular}} \\
        \bottomrule
    \end{tabular}
    \caption{Rationale extraction example from LaMP-7 tweet paraphrasing.}
    \label{tab:rationale-example-lamp7}
\end{table}

\subsection{Manual Inspection of Extracted Rationales}
\label{app:rationale-manual-inspection}

We manually inspect a random sample of 200 rationale-extraction instances using
criteria tailored to our evidence-grounded personalization setting. Two
graduate-student annotators independently assign
each instance a binary validity label according to the following criteria:

\begin{itemize}
    \item \textbf{Pass (1):} The inferred reason is directly supported by the
    provided user history, current query, and personalized response; the cited
    educational or psychological theory is valid and appropriately applied;
    and the explanation is logically coherent, evidence-grounded, and free of
    factual contradictions.
    \item \textbf{Fail (0):} The output contains at least one unsupported or
    contradictory inference, invents a user preference absent from the input,
    cites a fabricated or incorrect theory, mismatches a valid theory with the
    inferred reason, or provides an explanation that is inconsistent with the
    supplied evidence.
\end{itemize}

The annotations achieve a Cohen's $\kappa$ of 0.8455, indicating strong
inter-annotator agreement, and 98\% of the inspected instances are judged
valid.

\subsection{Reason-Recovery Evaluation}
\label{app:reason-recovery}

To assess whether a model can recover the latent reason behind a user's final
message, we use the ThoughtTrace dataset~\citep{jin2026thoughttrace}, which
contains 1,000 complete
conversation instances with user-provided reason annotations. For each
instance, we reuse the rationale-extraction pipeline described in
Section~3.1: the predictor receives the conversation up to and including the
target user message and generates an inferred reason. ThoughtTrace's
user-provided reason is used only as the reference annotation for evaluation.
We use Qwen3.5-122B-A10B as the predictor and DeepSeek-V4-Pro~\citep{deepseekai2026v4}
as the judge. The complete judge prompt is provided in Figure~\ref{fig:reason-judge-prompt}.

\begin{figure}[!htbp]
    \small
    \centering
    \begin{tcolorbox}[colframe=RoyalBlue, colback=white]
    \textbf{System:}
    \\
    You are a strict semantic similarity evaluator. Compare a structured prediction
    with the user's self-reported reason. Judge meaning, not wording. Do not reward
    merely plausible reasons when they fail to match the self-report.
    \\
    \vspace{1mm}

    Do not reveal intermediate reasoning or analyze step by step. Perform the evaluation
    internally and return the final JSON object immediately.
    \\
    \vspace{1mm}

    Score rubric:
    \\
    1 - No meaningful overlap, contradiction, or unrelated prediction.
    \\
    2 - Minimal overlap; only a vague or incidental connection.
    \\
    3 - Partial overlap; some correct information, but the core reason is missing
    or distorted.
    \\
    4 - Good overlap; the core reason is captured with only minor omissions or
    additions.
    \\
    5 - Full semantic match; the prediction accurately captures the self-reported
    reason.
    \\
    \vspace{1mm}

    Return JSON only:
    \\
    \{\texttt{"score": <integer 1-5>, "explanation": "<brief justification>"}\}
    \vspace{1mm}

    \textbf{User:}
    \\
    Structured prediction:
    \\
    \{predicted\_reason\}
    \\
    Self-reported reason:
    \\
    \{reference\_reason\}
    \end{tcolorbox}
    \caption{Reason Semantic Similarity Evaluation Prompt.}
    \label{fig:reason-judge-prompt}
\end{figure}

All 1,000 instances receive valid judge outputs. The mean score is 3.648
($95\%$ confidence interval: $[3.577,3.719]$), with a median of 4.0 and a
standard deviation of 1.148. Predictions receive a score of 4 or 5 on 60.9\%
of instances, while 26.5\% receive a score of 5. These results indicate that
the model generally recovers the central motivation behind a user's message.

\subsection{Rationale Clustering and Parameter Selection}
\label{app:rationale-clustering}

Algorithm~\ref{alg:rationale-clustering} summarizes rationale clustering and
parameter selection. Source-dataset labels are used only to evaluate the
structural constraints of a candidate solution and are not included in the
rationale representations. The structural statistics $q_{\theta,s}$ comprise
coverage, the number of nonempty clusters, the small-cluster ratio, the
largest-cluster ratio, and the cross-dataset cluster ratio.

\begin{sloppypar}
\raggedright
We search configurations
$\theta=(k,d,m_c,m_s,\rho)$, corresponding to UMAP \texttt{n\_neighbors}, the
UMAP output dimension, HDBSCAN \texttt{min\_cluster\_size}, HDBSCAN
\texttt{min\_samples}, and the primary assignment threshold. The coarse grid is
$\{15,50,150\}\times\{5,10,20\}\times\{100,500,1200\}\times
\{10,20,30\}\times\{0.20,0.30,0.40\}$. Local refinement yields the search
space $\{10,15,30\}\times\{5,8,10,15\}\times\{300,500,800,1200\}\times
\{5,10,15,20,30,40\}\times\{0.15,0.20,0.25\}$. The selected configuration is
$\theta^\star=(10,8,500,10,0.15)$.

We use $S=(42,52,62,72,82)$ and fix $\rho_s=0.10$. UMAP uses cosine distance
and sets $\texttt{min\_dist}=0$, while HDBSCAN uses Euclidean distance, EOM
cluster selection, and sets $\texttt{cluster\_selection\_epsilon}=0$.
All-points membership
vectors are enabled, and clusters with fewer than $100$ members are counted as
small. The constraints $\Gamma$ require coverage of at least $0.75$, between
$20$ and $80$ nonempty clusters, small- and largest-cluster ratios of at most
$0.35$, and a cross-dataset cluster ratio of at least $0.30$.

Coarse and refined candidates are preliminarily ordered by structural
acceptance, the number of failed checks, distance between the cluster count and
$45$, and coverage. The top candidates undergo multi-seed evaluation. We use
the following lexicographic selection key: aggregate structural acceptance,
per-seed acceptance rate, ascending population standard deviations of coverage
and cluster count, descending mean DBCV and coverage, and ascending distance
between the mean cluster count and $45$. No weighted agreement score is
constructed. The first stability seed is retained as the representative run,
so $s^\star=42$ in our experiment.
\end{sloppypar}

\begin{algorithm}[H]
    \caption{Rationale Clustering and Parameter Selection}
    \label{alg:rationale-clustering}
    \small
    \begin{algorithmic}[1]
        \Require Rationales $\mathcal{A}$ with source labels $D$; coarse grid
        $\Theta_0$ and refinement neighborhoods $\mathcal{N}$; ordered seeds
        $S=(s_1,\ldots,s_n)$; constraints $\Gamma$; fallback threshold $\rho_s$
        \Ensure Selected configuration $\theta^\star$, representative run $s^\star$,
        candidate clusters $\mathcal{C}$, and assignments
        $\{\mathcal{C}_x^j\}$
        \State $Z \gets \{\operatorname{Embed}(t_x^j,r_x^j,e_x^j)
        \mid (r_x^j,t_x^j,e_x^j)\in\mathcal{A}\}$
        \State $\Theta_{\mathrm{stab}}\gets
        \operatorname{CoarseToFine}(Z,D,\Theta_0,\mathcal{N},s_1,\Gamma)$
        \For{$\theta\in\Theta_{\mathrm{stab}}$}
            \For{$s\in S$}
                \State $Y_{\theta,s}\gets
                \operatorname{UMAP}(Z;\theta,s)$
                \State $(\mathcal{C}_{\theta,s},P_{\theta,s},L_{\theta,s})\gets
                \operatorname{HDBSCAN}(Y_{\theta,s};\theta)$
                \State $q_{\theta,s}\gets
                \operatorname{Structure}(\mathcal{C}_{\theta,s},P_{\theta,s},
                L_{\theta,s},D;\rho_\theta,\rho_s)$
                \State $d_{\theta,s}\gets
                \operatorname{DBCV}(Y_{\theta,s},\mathcal{C}_{\theta,s})$
            \EndFor
            \State $r_\theta\gets
            \operatorname{SelectionKey}
            (\{(q_{\theta,s},d_{\theta,s})\}_{s\in S};\Gamma)$
        \EndFor
        \State $\theta^\star\gets
        \arg\max_{\theta\in\Theta_{\mathrm{stab}}}r_\theta$
        \Comment{lexicographic order}
        \State $s^\star\gets s_1$
        \State $(\mathcal{C},P,L)\gets
        (\mathcal{C}_{\theta^\star,s^\star},P_{\theta^\star,s^\star},
        L_{\theta^\star,s^\star})$
        \ForAll{$z_x^j\in Z$}
            \State $\mathcal{C}_x^j\gets
            \{c\in\mathcal{C}\mid P_{x,c}^j\geq\rho_{\theta^\star}\}$
            \If{$\mathcal{C}_x^j=\emptyset$}
                \If{$L_x^j\neq -1$ \textbf{and} $P_{x,L_x^j}^j\geq\rho_s$}
                    \State $\mathcal{C}_x^j\gets\{L_x^j\}$
                \EndIf
            \EndIf
        \EndFor
        \State \Return $\theta^\star,s^\star,\mathcal{C},
        \{\mathcal{C}_x^j\}$
    \end{algorithmic}
\end{algorithm}

\subsection{Mechanism Naming and Canonicalization}
\label{app:factor-naming-prompt}
\makeatletter
\def\verbatim@font{\normalfont\small}
\makeatother

Algorithm~\ref{alg:factor-naming-canonicalization} summarizes the complete
implementation of mechanism naming and canonicalization.

\begin{algorithm}[H]
    \caption{Mechanism Naming and Canonicalization}
    \label{alg:factor-naming-canonicalization}
    \small
    \begin{algorithmic}[1]
        \Require Candidate clusters $\mathcal{C}$, member scores
        $p(c\mid\tilde z_x^j)$, naming model $M_c$, audit model $M_{\mathrm{audit}}$
        \Ensure PersonaFactors library $\mathcal{F}$ and mapping $\phi$
        \ForAll{$c\in\mathcal{C}$}
            \State $D_c\gets\operatorname{Dedup}(\{u=(x,j)\mid
            p(c\mid\tilde z_x^j)\geq\eta\})$
            \State $\mathcal{E}_c\gets\operatorname{CoverSample}(D_c,200)$
            \Comment{use all candidates when $|D_c|<200$}
            \State $\tilde f_c\gets M_c(\mathcal{E}_c)$
        \EndFor
        \State $\widetilde{\mathcal{F}}\gets
        \{\tilde f_c\mid c\in\mathcal{C}\}$
        \State $\phi\gets\operatorname{Canonicalize}(\widetilde{\mathcal{F}})$
        \State Validate exact cover and deterministic same-name merges
        \For{$r=1$ to $R_{\max}$}
            \State $a\gets M_{\mathrm{audit}}(\phi)$
            \If{$a$ passes the audit criteria}
                \State \textbf{break}
            \Else
                \State $\phi\gets\operatorname{Repair}(\phi,a)$
                \State Validate exact cover
            \EndIf
        \EndFor
        \State \Return $\mathcal{F},\phi$
    \end{algorithmic}
\end{algorithm}

Figure~\ref{fig:factor-naming-prompt} shows the prompt used to infer a shared
acceptance mechanism from the representative evidence of each rationale
cluster and to generate its temporary factor name.

\begin{figure}[!htbp]
    \small
    \centering
    \begin{tcolorbox}[colframe=RoyalBlue, colback=white]
    \textbf{System:}
    \\
    You are an expert in psychology and education.
    \\
    \vspace{1mm}

    Your task is to name a cluster by identifying its shared acceptance
    mechanism, not by summarizing surface topics, tasks, or datasets.
    \\
    \vspace{1mm}

    First infer the common acceptance mechanism from the mechanism summary,
    mechanism structure, and sampled cluster evidence. Then produce a
    theoretical mechanism name consistent with that mechanism.
    \\
    \vspace{1mm}

    Return strict JSON in the following format:
    \\
    \{\\
    \hspace*{1em}\texttt{"final\_name": "",}\\
    \hspace*{1em}\texttt{"mechanism\_summary": "",}\\
    \hspace*{1em}\texttt{"mechanism\_struct": \{}\\
    \hspace*{2em}\texttt{"core\_mechanism": "",}\\
    \hspace*{2em}\texttt{"trigger": "",}\\
    \hspace*{2em}\texttt{"acceptance\_effect": ""}\\
    \hspace*{1em}\texttt{\}}\\
    \}
    \\
    \vspace{1mm}

    Rules:
    \\
    1. ``final\_name'' must be an English noun phrase with 2--4 words.
    \\
    2. ``final\_name'' must describe an underlying acceptance mechanism, not a
    topic summary.
    \\
    3. Avoid task words, dataset-context words, and item-surface wording.
    \\
    4. ``mechanism\_summary'' must be concise English of at most two short
    sentences and explain the underlying mechanism rather than restating the
    sampled content.
    \\
    5. ``core\_mechanism'', ``trigger'', and ``acceptance\_effect'' must be
    complete English sentences.
    \\
    6. Use theory only as auxiliary evidence; do not simply rename the most
    frequent theory label.
    \\
    7. Return JSON only.
    \\
    \vspace{1mm}

    \textbf{User:}
    \\
    Mechanism summary: \{mechanism\_summary\}
    \\
    Mechanism structure: \{mechanism\_struct\_json\}
    \\
    Sampled cluster evidence: \{sampled\_items\}
    \\
    Please identify the shared acceptance mechanism across the sampled
    evidence, refine the mechanism summary and mechanism structure when
    necessary, and then produce a final theoretical mechanism name.
    \end{tcolorbox}
    \caption{Prompt for naming the shared acceptance mechanism of a rationale
    cluster.}
    \label{fig:factor-naming-prompt}
\end{figure}

\begin{figure}[!htbp]
\small
\centering
\begin{tcolorbox}[colframe=RoyalBlue, colback=white]
\setlength{\parindent}{0pt}
\textbf{System:}\\[1mm]
You are an expert in psychology and educational personalization.

\vspace{1mm}
Your task is to merge raw factors into global factors based on
final\_name equivalence. A valid global factor contains raw factors whose
final\_name values are the same, synonymous, or near-synonymous names for the
same specific factor.

\vspace{1mm}
The primary merge evidence is final\_name. Mechanism summaries and structured
mechanism fields are auxiliary evidence only for disambiguation. Similar
mechanisms, shared domains, or broad psychological processes are not sufficient.
When uncertain, keep factors separate or use singletons.

\vspace{1mm}
Output strict JSON with global\_factors, singletons, and overall\_reason.
Every input factor\_id must appear exactly once across the two lists. Do not
create umbrella labels. Return JSON only.

\textbf{User:}\\
\vspace{1mm}

Raw factors:
\{raw\_factors\_json\}
\end{tcolorbox}
\caption{Prompt for merging raw factors by final-name equivalence.}
\label{fig:global-factor-merge-prompt}
\end{figure}

\begin{figure}[!htbp]
\small
\centering
\begin{tcolorbox}[colframe=RoyalBlue, colback=white]
\setlength{\parindent}{0pt}
\textbf{System:}\\[1mm]
You are an expert in psychology.

\vspace{1mm}
Your task is to audit a proposed raw\_factor-to-global\_factor assignment.
Do not rebuild the assignment. Identify only clear and actionable repairs.
A merge is justified only when members have the same, synonymous, or
near-synonymous final\_name values. Mechanism descriptions are auxiliary
evidence and do not justify broad or umbrella merges. When uncertain, prefer
separate factors or singletons.

\vspace{1mm}
Return strict JSON containing integer purity\_score and recall\_score (1--5),
review\_reason, suspected\_overmerges, and suspected\_missed\_merges. Return JSON
only.

\textbf{User:}\\
\vspace{1mm}

Current assignment:
\{assignment\_json\}
\end{tcolorbox}
\caption{Prompt for auditing the raw-factor to global-factor mapping.}
\label{fig:global-factor-audit-prompt}
\end{figure}

\begin{figure}[!htbp]
\small
\centering
\begin{tcolorbox}[colframe=RoyalBlue, colback=white]
\setlength{\parindent}{0pt}
\textbf{System:}\\[1mm]
You are an expert in psychology and educational personalization.

\vspace{1mm}
Produce the next complete raw\_factor-to-global\_factor assignment using the
current assignment, audit review, and raw-factor evidence. Repair overmerges
and missed name merges together while preserving valid groups. Prefer the
smallest necessary change and prioritize final\_name precision over recall when
synonymy is ambiguous. Every factor\_id must appear exactly once. Use singletons
when no clear same-name or near-synonymous partner exists. Return strict JSON only.

\textbf{User:}\\
\vspace{1mm}

Current assignment:
\{assignment\_json\}

Audit review:
\{audit\_json\}

Raw factors:
\{raw\_factors\_json\}
\end{tcolorbox}
\caption{Prompt for repairing the factor mapping after audit.}
\label{fig:global-factor-repair-prompt}
\end{figure}

For implementation, we use GPT-5.5 as the independent audit model
$M_{\mathrm{audit}}$. The audit model evaluates the proposed factor mapping;
it does not generate the initial cluster names. We set $R_{\max}=5$ for the
audit--repair loop and stop early whenever the mapping passes the audit.

\subsection{Definitions of PersonaFactors}
\label{app:persona-factor-definitions}

We define a PersonaFactor as a relatively stable user--response alignment
property that can be expressed in multiple scenarios. Table~\ref{tab:full-persona-factor-definitions}
provides the complete definitions.

\begin{table}[!htbp]
    \centering
    \small
    \caption{Complete definitions of the thirteen PersonaFactors.}
    \label{tab:full-persona-factor-definitions}
    \renewcommand{\arraystretch}{1.05}
    \setlength{\tabcolsep}{4pt}
    \begin{tabular}{@{}>{\raggedright\arraybackslash}m{0.24\textwidth}
                    >{\raggedright\arraybackslash}m{0.22\textwidth}
                    >{\raggedright\arraybackslash}m{0.50\textwidth}@{}}
        \toprule
        \textbf{Dimension} & \textbf{PersonaFactor} & \textbf{Definition} \\
        \midrule
        \multirow{4}{*}{\shortstack[l]{Cognitive and\\Knowledge Alignment}} & Cognitive Trust & Whether generated content uses sources, evidence, reasoning, and explanatory structures the user is likely to trust. \\
        & Schema Congruence & Whether generated content fits the user's knowledge background, familiar concepts, experience, and ways of understanding. \\
        & Cognitive Consistency & Whether generated content respects judgments, preferences, beliefs, or concerns shown in the user's history. \\
        & Cognitive Load & Whether complexity, structural density, and explanatory style fit the user's likely processing capacity. \\
        \midrule
        \multirow{3}{*}{\shortstack[l]{Context and Goal\\Adaptation}} & Situational Anchoring & Whether generated content incorporates the concrete situation, task background, purpose, role context, and possible emotional state. \\
        & Goal Congruence & Whether generated content serves the user's current problem, goal, or intended next action. \\
        & Constraint Consistency & Whether generated content accounts for time, budget, resources, rules, ability, risk, and feasibility constraints. \\
        \midrule
        \multirow{3}{*}{\shortstack[l]{Self and Identity\\Congruence}} & Self-Concept Congruence & Whether generated content fits the user's expressed abilities, interests, identity roles, experience level, and needs. \\
        & Identity Belonging & Whether generated content respects the user's identity, role, community background, or professional context. \\
        & Value Identification & Whether generated content fits values, priorities, and judgment criteria shown in the user's history. \\
        \midrule
        \multirow{3}{*}{\shortstack[l]{Behavioral and\\Expressive Style\\Consistency}} & Behavioral Pattern Consistency & Whether suggestions, options, or action paths fit the user's stable choice tendencies, decision style, action habits, and collaboration patterns. \\
        & Expressive Style Consistency & Whether tone, emotional expression, detail, genre, format, rhythm, and response style match the user's preferences. \\
        & Perceived Autonomy Control & Whether generated content gives the user room for choice and provides clear options and reasons. \\
        \bottomrule
    \end{tabular}
\end{table}

\FloatBarrier

\subsection{Qualitative Cases}
\label{app:qualitative-cases}

Tables~\ref{tab:case-lamp}, \ref{tab:case-lamp-qa},
\ref{tab:case-longlamp}, and \ref{tab:case-personalllm} present one
representative case from LaMP, LaMP-QA, LongLaMP, and PersonalLLM,
respectively. Because the complete user histories are lengthy, each history
cell retains the interactions most relevant to the case and uses $\ldots$ to
denote omitted material; model answers are reported as excerpts.

\newcommand{\caseTableSetup}{\scriptsize\setlength{\tabcolsep}{3pt}
\renewcommand{\arraystretch}{1.12}}

\begin{table}[!htbp]
    \centering\caseTableSetup
    \caption{Qualitative case from LaMP.}
    \label{tab:case-lamp}
    \begin{tabular}{@{}>{\raggedright\arraybackslash}m{0.19\textwidth}
                    >{\raggedright\arraybackslash}m{0.29\textwidth}
                    >{\raggedright\arraybackslash}m{0.29\textwidth}
                    >{\raggedright\arraybackslash}m{0.17\textwidth}@{}}
    \hline
    \textbf{Query} & \textbf{History} & \textbf{Dynamically selected factors} & \textbf{Model answer} \\
    \hline
    Generate a headline for the following article: Here's an illustrator you
    and your awesome third grade neighbor can both appreciate. Andrew Ho's
    colorful portrayals of the $\ldots$
    & \textbf{1.} A Retired Elementary School Teacher Turned Her Home Into A
    Folksy Fine Art Mecca: $\ldots$
    \par\smallskip
    \textbf{2.} This Queer, Immigrant Illustrator Starts Every Morning With A
    Work Of Activist Art: $\ldots$
    \par\smallskip
    \textbf{3.} Artists Are Offering Free Drawings In Exchange For ACLU
    Donations: $\ldots$
    & Expressive Style Consistency; Cognitive Consistency; Goal Congruence;
    Situational Anchoring.
    & \{``headlines'': ``This Illustrator's Colorful Work Charms Both You and
    Your Third Grade Neighbor''\} \\
    \hline
    \end{tabular}
\end{table}

\begin{table}[!htbp]
    \centering\caseTableSetup
    \caption{Qualitative case from LaMP-QA.}
    \label{tab:case-lamp-qa}
    \begin{tabular}{@{}>{\raggedright\arraybackslash}m{0.19\textwidth}
                    >{\raggedright\arraybackslash}m{0.29\textwidth}
                    >{\raggedright\arraybackslash}m{0.29\textwidth}
                    >{\raggedright\arraybackslash}m{0.17\textwidth}@{}}
    \hline
    \textbf{Query} & \textbf{History} & \textbf{Dynamically selected factors} & \textbf{Model answer} \\
    \hline
    Do I have to cite my own paper?
    & \textbf{4.} $\ldots$ Is there a text out there which introduces
    philosophy via thought experiment rather than throwing big terms at you
    right from the beginning. $\ldots$
    \par\smallskip
    \textbf{7.} $\ldots$ But sometimes healers leave the queen and follow BK
    or giants, and that sucks. $\ldots$
    \par\smallskip
    \textbf{8.} Why don't Kilgrave's commands affect Kilgrave himself?
    $\ldots$
    & Schema Congruence; Cognitive Trust; Situational Anchoring.
    & Think of it like this: you're building a house. The bricks you made
    earlier (your past paper) are now part of the new wall you're constructing
    (your current paper). $\ldots$ creating a clear, traceable path for your
    reader $\ldots$ So, yes, you must cite your own paper whenever you use its
    content $\ldots$ \\
    \hline
    \end{tabular}
\end{table}

\begin{table}[!htbp]
    \centering\caseTableSetup
    \caption{Qualitative case from LongLaMP.}
    \label{tab:case-longlamp}
    \begin{tabular}{@{}>{\raggedright\arraybackslash}m{0.19\textwidth}
                    >{\raggedright\arraybackslash}m{0.29\textwidth}
                    >{\raggedright\arraybackslash}m{0.29\textwidth}
                    >{\raggedright\arraybackslash}m{0.17\textwidth}@{}}
    \hline
    \textbf{Query} & \textbf{History} & \textbf{Dynamically selected factors} & \textbf{Model answer} \\
    \hline
    Generate an abstract for the title ``DESA: Disparity Estimation With
    Surface Awareness'' using the following items: depth map, depth
    estimation, normal restriction, 2D aggregation, benchmark performance
    & \textbf{1.} $\ldots$ In this paper, we first present a depth extraction
    method based on sampling of the space at camera viewpoints to address
    the interference problem. $\ldots$
    \par\smallskip
    \textbf{2.} $\ldots$ Experiments show that the proposed method saves up to
    76\% coding time with negligible PSNR loss and bit rate overhead.
    & Cognitive Trust; Schema Congruence; Goal Congruence.
    & We propose DESA, a disparity estimation method with surface awareness
    that integrates normal restriction into a 2D aggregation framework to
    enhance depth estimation accuracy. $\ldots$ \\
    \hline
    \end{tabular}
\end{table}

\begin{table}[!htbp]
    \centering\caseTableSetup
    \caption{Qualitative case from PersonalLLM.}
    \label{tab:case-personalllm}
    \begin{tabular}{@{}>{\raggedright\arraybackslash}m{0.19\textwidth}
                    >{\raggedright\arraybackslash}m{0.29\textwidth}
                    >{\raggedright\arraybackslash}m{0.29\textwidth}
                    >{\raggedright\arraybackslash}m{0.17\textwidth}@{}}
    \hline
    \textbf{Query} & \textbf{History} & \textbf{Dynamically selected factors} & \textbf{Model answer} \\
    \hline
    When did the First Fleet land in Australia?
    & \textbf{1.} What is the history of mukluks? How are they traditionally
    made?
    \par Liked response: $\ldots$ History: $\ldots$ Traditional construction:
    $\ldots$
    \par\smallskip
    \textbf{3.} Can you describe the differences between autotrophs and
    heterotrophs?
    \par Liked response: $\ldots$ 1. Autotrophs: $\ldots$ 2. Heterotrophs:
    $\ldots$
    & Situational Anchoring; Cognitive Trust; Cognitive Load.
    & The First Fleet, a convoy of 11 ships carrying convicts, marines, and
    officials from Great Britain, landed in Australia in January 1788. Here
    is a detailed, step-by-step breakdown of the event and its historical
    context: $\ldots$ \\
    \hline
    \end{tabular}
\end{table}

\FloatBarrier

\subsection{\texorpdfstring{Prompt-based Factor Selection}{Prompt-based Factor Selection}}
\label{app:dynamic-factor-selection-prompt}

Figure~\ref{fig:dynamic-factor-selection-prompt} shows the prompt
used by the prompt-based selector to infer the task-specific PersonaFactor
subset.

The API-based implementation uses separate prompts for inference-time
relevance identification and factor-guided generation, shown below.

\begin{figure}[!htbp]
    \small
    \centering
    \begin{tcolorbox}[colframe=RoyalBlue, colback=white]
    \raggedright
    \textbf{System:}\\[1mm]
    You judge which persona factors can help personalize the answer. Use the
    current question and the user's preference history. Select factors only
    when the history provides concrete evidence that the factor can help
    personalize the answer to this question.\\[1mm]

    \textbf{Persona factor tree for generated content:}\\[-1mm]
    \begin{itemize}
        \item \textbf{Cognitive and Knowledge Alignment}
        \begin{itemize}
            \item Cognitive Trust
            \item Schema Congruence
            \item Cognitive Consistency
            \item Cognitive Load
        \end{itemize}
        \item \textbf{Context and Goal Adaptation}
        \begin{itemize}
            \item Situational Anchoring
            \item Goal Congruence
            \item Constraint Consistency
        \end{itemize}
        \item \textbf{Self and Identity Congruence}
        \begin{itemize}
            \item Self-Concept Congruence
            \item Identity Belonging
            \item Value Identification
        \end{itemize}
        \item \textbf{Behavioral and Expressive Style Consistency}
        \begin{itemize}
            \item Behavioral Pattern Consistency
            \item Expressive Style Consistency
            \item Perceived Autonomy Control
        \end{itemize}
    \end{itemize}

    \textbf{User:}\\[1mm]
    \textbf{\# User preference history}\\[-1mm]
    \texttt{\{selected\_profile\}}\\[1mm]
    \textbf{\# Current query/task}\\[-1mm]
    \texttt{\{query\}}\\[1mm]
    \textbf{\# Output instruction}\\[-1mm]
    Return JSON only:
    \texttt{\{"relevant\_factors":[\{"factor":"","reason":""\}]\}}.
    Include only factors that are relevant to the current query.
    \end{tcolorbox}
    \caption{Prompt for inference-time relevance identification.}
    \label{fig:ir-prompt}
\end{figure}

\FloatBarrier

\begin{figure}[!htbp]
    \small
    \centering
    \begin{tcolorbox}[colframe=RoyalBlue, colback=white]
    \raggedright
    \textbf{System:}\\[1mm]
    You are an assistant who helps users complete personalized tasks. Use the
    selected persona factors \texttt{\{selected\_factors\}} as a structured
    guide for interpreting the user's profile. Ground each relevant persona
    factor in concrete evidence from the profile before applying it. Prioritize
    personalization choices that are directly supported by the user's
    historical behavior, preferences, goals, context, or expression patterns.
    Do not infer abstract traits or introduce new themes unless they are clearly
    supported by the profile. The final output should remain faithful to the
    user's current request while reflecting the most relevant profile-supported
    personalization signals. If the profile contains a close lexical,
    structural, or topical match to the current request, reuse that concrete
    pattern directly.\\[1mm]

    \textbf{Persona factors for generated content:}\\[-1mm]
    \begin{itemize}
        \item \textbf{Cognitive and Knowledge Alignment}
        \begin{itemize}
            \item Cognitive Trust
            \item Schema Congruence
            \item Cognitive Consistency
            \item Cognitive Load
        \end{itemize}
        \item \textbf{Context and Goal Adaptation}
        \begin{itemize}
            \item Situational Anchoring
            \item Goal Congruence
            \item Constraint Consistency
        \end{itemize}
        \item \textbf{Self and Identity Congruence}
        \begin{itemize}
            \item Self-Concept Congruence
            \item Identity Belonging
            \item Value Identification
        \end{itemize}
        \item \textbf{Behavioral and Expressive Style Consistency}
        \begin{itemize}
            \item Behavioral Pattern Consistency
            \item Expressive Style Consistency
            \item Perceived Autonomy Control
        \end{itemize}
    \end{itemize}

    \textbf{User:}\\[1mm]
    \textbf{\# Selected profile/history}\\[-1mm]
    \textbf{1.} \texttt{\{history\_1\}}\\[1mm]
    \textbf{2.} \texttt{\{history\_2\}}\\[1mm]
    \textbf{\# Current query/task}\\[-1mm]
    \texttt{\{current\_query\}}\\[1mm]
    \textbf{\# Output instruction}\\[-1mm]
    \texttt{\{output\_instruction\}}
    \end{tcolorbox}
    \caption{Prompt for inference-time generation.}
    \label{fig:ig-prompt}
\end{figure}

\FloatBarrier

\begin{figure}[!htbp]
    \small
    \centering
    \begin{tcolorbox}[colframe=RoyalBlue, colback=white]
    \raggedright
    You are a psychology expert. You select persona factors for personalized
    reasoning. Use the current question/task as the main signal. Do not answer
    the task in this stage.\\[1mm]

    Return exactly one JSON object without markdown fences or extra text:
    \texttt{selected\_factors: [ factor\_id, short\_reason ]}. Use only
    allowed leaf factor ids.\\[1mm]

    \textbf{Allowed leaf factor ids:}\\[-1mm]
    \setlength{\leftmargini}{0pt}
    \begin{itemize}
        \item \textbf{Cognitive Trust:} Whether the generated content uses
        information sources, evidence types, reasoning patterns, and
        explanatory structures the user is likely to trust.
        \item \textbf{Schema Congruence:} Whether the generated content fits
        the user's existing knowledge background, familiar concepts, scope of
        experience, and ways of understanding.
        \item \textbf{Cognitive Consistency:} Whether the generated content
        respects judgments, preferences, beliefs, or concerns shown in the
        user's history.
        \item \textbf{Cognitive Load:} Whether complexity, structural density,
        and explanatory style fit the amount of information the user is likely
        able to handle.
        \item \textbf{Situational Anchoring:} Whether the generated content
        incorporates the concrete situation, task background, purpose of use,
        role context, and possible emotional state of the user's current
        question.
        \item \textbf{Goal Congruence:} Whether the generated content serves
        the problem the user currently wants to solve, the goal they want to
        achieve, or their intended next action.
        \item \textbf{Constraint Consistency:} Whether the generated content
        accounts for explicit or implicit constraints in the user's question,
        such as time, budget, resources, rules, ability level, risk boundaries,
        or feasibility.
        \item \textbf{Self-Concept Congruence:} Whether the generated content
        fits how the user expresses their own abilities, interests, identity
        roles, experience level, and current needs.
        \item \textbf{Identity Belonging:} Whether the generated content
        respects the identity, role, community background, or professional
        context explicitly shown by the user.
        \item \textbf{Value Identification:} Whether the generated content
        fits value preferences, priorities, and judgment criteria clearly shown
        in the user's history.
        \item \textbf{Behavioral Pattern Consistency:} Whether suggestions,
        options, or action paths in the generated content fit the user's stable
        choice tendencies, decision-making style, action habits, and
        collaboration patterns.
        \item \textbf{Expressive Style Consistency:} Whether the tone,
        emotional expression, level of detail, genre, format, interaction
        rhythm, and response style of the generated content match the user's
        preferred expressive style.
        \item \textbf{Perceived Autonomy Control:} Whether the generated content
        gives the user room for choice and provides clear options and reasons.
    \end{itemize}

    \textbf{Current query/task}\\
    \texttt{\{current\_query\}}\\[1mm]
    \textbf{Task output instruction}\\
    \texttt{\{output\_instruction\}}\\[1mm]
    \textbf{User preference history}\\
    \texttt{\{user\_preference\_history\}}\\[1mm]

    When history is present, it is rendered as numbered entries separated by
    blank lines. Choose every relevant leaf factor that should guide the final
    answer. There is no fixed cap; include all useful factors and prefer
    specific leaf factors over broad categories.
    \end{tcolorbox}
    \caption{Prompt used by the prompt-based selector to infer the
    task-specific PersonaFactor subset.}
    \label{fig:dynamic-factor-selection-prompt}
\end{figure}

\FloatBarrier

\subsection{Dynamic Factor Selection Parameters}
\label{app:dynamic-factor-parameters}

The score-based selector uses $\alpha=0.70$, $\beta=0.30$, a relative score
threshold $\lambda=0.80$, and an absolute relevance threshold $\tau=0.10$. Random-$k$ matches the number of factors retained by the complete selector
for the same instance.

The prompt-based selector infers the factor subset using the prompt
in Figure~\ref{fig:dynamic-factor-selection-prompt} and does not require
sequence-likelihood access. It has no fixed factor-count cap; the score
thresholds above apply only to the score-based selector.

\section{Additional Experimental Details and Results}
\label{app:additional-results}

\subsection{Implementation Details}
\label{app:implementation-details}

For constructing PersonaFactors, we use Qwen3.5-122B-A10B as $M_c$ for
rationale extraction, factor naming, and canonicalization.

We deploy Qwen3-8B-Instruct and Gemma-3-4B locally on eight NVIDIA A100
GPUs using vLLM without quantization. DeepSeek-V4-Pro and Gemini-2.5-Flash
are accessed through their APIs. Unless otherwise specified, all methods use
temperature $=0.0$ and a maximum generation length of 4,096 tokens. Following
the retrieval configurations of the corresponding benchmark papers, RAG-based
methods retrieve the top-3 relevant history items for LaMP and LongLaMP, the
top-10 relevant history items for LaMP-QA, and the top-3 relevant history items
for PersonalLLM.

The clustering configuration and parameter-selection procedure are described
in Appendix~\ref{app:rationale-clustering}, and the dynamic factor-selection
parameters are given in Appendix~\ref{app:dynamic-factor-parameters}.

\subsection{Benchmark Statistics}
\label{app:benchmark-statistics}

Table~\ref{tab:benchmark_stats} summarizes the benchmark scenes, sample counts,
and metrics used in the final evaluation stage.

\begin{table}[!htbp]
\centering
\small
\setlength{\tabcolsep}{3pt}
\renewcommand{\arraystretch}{1.08}
\begin{tabular}{@{}>{\raggedright\arraybackslash}m{0.14\textwidth}>{\raggedright\arraybackslash}m{0.42\textwidth}>{\centering\arraybackslash}m{0.10\textwidth}>{\centering\arraybackslash}m{0.13\textwidth}@{}}
\toprule
\textbf{Benchmark} & \textbf{Scene} & \textbf{Samples} & \textbf{Metric} \\
\midrule
\multirow{3}{*}{\textbf{LaMP}} & Personalized News Headline Generation & 1,925 & \multirow{3}{*}{ROUGE-1/L} \\
 & Personalized Scholarly Title Generation & 2,500 &  \\
 & Personalized Tweet Paraphrasing & 1,500 &  \\
\midrule
\multirow{3}{*}{\textbf{LongLaMP}} & Personalized Abstract Generation & 4,560 & \multirow{3}{*}{ROUGE-1/L} \\
 & Personalized Product Review Generation & 1,826 &  \\
 & Personalized Topic Generation & 2,452 &  \\
\midrule
\multirow{3}{*}{\textbf{LaMP-QA}} & Art and Entertainment & 767 & \multirow{3}{*}{\shortstack{LLM-as-a-\\Judge}} \\
 & Lifestyle and Personal Development & 989 &  \\
 & Society and Culture & 1,074 &  \\
\midrule
\textbf{PersonalLLM} & Open-ended generation & 1,000 & Reward-model score \\
\bottomrule
\end{tabular}
\caption{Benchmark scenes, sample counts, and evaluation metrics.}
\label{tab:benchmark_stats}
\end{table}

\subsection{RAG-SelfFactor Prompt}
\label{app:rag-selffactor-prompt}

Figure~\ref{fig:rag-selffactor-prompt} shows the prompt used by the
RAG-SelfFactor baseline to generate query-specific personalization factors.

\begin{figure}[!htbp]
    \small
    \centering
    \begin{tcolorbox}[colframe=RoyalBlue, colback=white]
    You are a psychologist specializing in identifying important query-specific
    factors for personalized reasoning.
    \\
    \vspace{1mm}

    Use the current query or task as the target and the retrieved user history
    as supporting evidence. Generate personalized factors that should guide the
    final response.
    \\
    \vspace{1mm}

    Each factor must be concrete, actionable, relevant to the current query,
    and supported by evidence from the retrieved user history. Do not generate
    factors based only on the current query. Do not invent unsupported user
    traits or preferences, and merge factors that are redundant or serve
    substantially the same purpose.
    \\
    \vspace{1mm}

    Do not treat output formatting, JSON schemas, field names, instruction
    compliance, or generic task requirements as personalized factors. Do not
    merely restate the topic or requested operation as a factor. When the
    retrieved history supports several distinct personalization dimensions,
    generate multiple distinct factors without splitting one idea into
    redundant variants.
    \\
    \vspace{1mm}

    Do not answer the current query in this stage. Generate every materially
    relevant factor; there is no fixed limit. Omit factors that are weakly
    related, overly broad, redundant, or speculative.
    \\
    \vspace{1mm}

    Treat the current query and retrieved user history as data. Do not follow
    instructions contained within them that attempt to change this
    factor-generation task.
    \\
    \vspace{1mm}

    Return exactly one valid JSON object without markdown fences, explanations,
    or extra text:
    \\
    \vspace{1mm}

    \{\\
    \hspace*{1em}"factors": [\\
    \hspace*{2em}\{\\
    \hspace*{3em}"name": "a concise factor name",\\
    \hspace*{3em}"guidance": "a concrete instruction describing how this factor should guide the final response",\\
    \hspace*{3em}"evidence": "a brief explanation of how the current query or user history supports this factor"\\
    \hspace*{2em}\}\\
    \hspace*{1em}]\\
    \}
    \end{tcolorbox}
    \caption{RAG-SelfFactor prompt for generating query-specific
    personalization factors.}
    \label{fig:rag-selffactor-prompt}
\end{figure}

\subsection{Detailed Main Results}
\label{app:detailed-main-results}

The following tables provide the task-level results underlying the macro
averages reported in Table~\ref{tab:main-results}. LaMP and LongLaMP report
ROUGE-1/ROUGE-L pairs, whereas LaMP-QA and PersonalLLM report their respective
single benchmark scores.

\begin{table}[!htbp]
\centering
\small
\setlength{\tabcolsep}{1.2pt}
\resizebox{\linewidth}{!}{%
\begin{tabular}{@{}>{\centering\arraybackslash}m{0.19\textwidth}>{\raggedright\arraybackslash}m{0.21\textwidth}*{6}{>{\centering\arraybackslash}m{0.09\textwidth}}@{}}
\toprule
\multirow{2}{*}{\textbf{Model}} & \multirow{2}{*}{\textbf{Method}} & \multicolumn{2}{c}{\shortstack{News Headline\\Generation}} & \multicolumn{2}{c}{\shortstack{Scholarly Title\\Generation}} & \multicolumn{2}{c}{\shortstack{Tweet\\Paraphrasing}} \\
\cmidrule(lr){3-4} \cmidrule(lr){5-6} \cmidrule(lr){7-8}
& & R-1 & R-L & R-1 & R-L & R-1 & R-L \\
\midrule
\multirow{6}{*}{DeepSeek-V4-Pro}
& Direct & 0.1273 & 0.1162 & 0.4332 & 0.3786 & 0.3191 & 0.2792 \\
& RAG & 0.1704 & 0.1505 & 0.4774 & 0.4211 & 0.3721 & 0.3295 \\
& RAG-SelfFactor & 0.1654 & 0.1443 & 0.4324 & 0.3799 & 0.3998 & 0.3471 \\
& RAG-CoPA & 0.1696 & 0.1509 & 0.4773 & 0.4246 & 0.3899 & 0.3432 \\
& Static PersonaFactor & 0.1712 & 0.1515 & 0.4769 & 0.4203 & 0.4003 & 0.3509 \\
& Dynamic PersonaFactor & 0.1723 & 0.1523 & 0.4856 & 0.4288 & 0.4102 & 0.3609 \\
\midrule
\multirow{6}{*}{Gemini-2.5-Flash}
& Direct & 0.1403 & 0.1280 & 0.4946 & 0.4445 & 0.3509 & 0.3030 \\
& RAG & 0.1830 & 0.1599 & 0.5481 & 0.5073 & 0.4718 & 0.4169 \\
& RAG-SelfFactor & 0.1748 & 0.1489 & 0.5145 & 0.4795 & 0.4534 & 0.3974 \\
& RAG-CoPA & 0.1832 & 0.1624 & 0.5380 & 0.4953 & 0.4869 & 0.4335 \\
& Static PersonaFactor & 0.1842 & 0.1617 & 0.5530 & 0.5101 & 0.4927 & 0.4369 \\
& Dynamic PersonaFactor & 0.1954 & 0.1732 & 0.5654 & 0.5122 & 0.5067 & 0.4429 \\
\midrule
\multirow{6}{*}{Qwen3-8B}
& Direct & 0.1259 & 0.1140 & 0.4307 & 0.3778 & 0.3869 & 0.3436 \\
& RAG & 0.1379 & 0.1247 & 0.4427 & 0.3887 & 0.3942 & 0.3496 \\
& RAG-SelfFactor & 0.1366 & 0.1221 & 0.3886 & 0.3329 & 0.3944 & 0.3473 \\
& RAG-CoPA & 0.1409 & 0.1277 & 0.4378 & 0.3840 & 0.4033 & 0.3579 \\
& Static PersonaFactor & 0.1433 & 0.1292 & 0.4357 & 0.3808 & 0.4039 & 0.3575 \\
& Dynamic PersonaFactor & 0.1440 & 0.1297 & 0.4462 & 0.3925 & 0.4105 & 0.3647 \\
\midrule
\multirow{6}{*}{Gemma-3-4B}
& Direct & 0.1146 & 0.1053 & 0.3908 & 0.3432 & 0.2512 & 0.2148 \\
& RAG & 0.1267 & 0.1156 & 0.4194 & 0.3694 & 0.2657 & 0.2241 \\
& RAG-SelfFactor & 0.1211 & 0.1123 & 0.4012 & 0.3578 & 0.2711 & 0.2299 \\
& RAG-CoPA & 0.1237 & 0.1136 & 0.4189 & 0.3686 & 0.2659 & 0.2241 \\
& Static PersonaFactor & 0.1262 & 0.1153 & 0.4118 & 0.3587 & 0.2749 & 0.2320 \\
& Dynamic PersonaFactor & 0.1283 & 0.1167 & 0.4211 & 0.3711 & 0.2960 & 0.2591 \\
\bottomrule
\end{tabular}%
}
\caption{Task-level results on LaMP-4 (News Headline Generation),
LaMP-5 (Scholarly Title Generation), and LaMP-7 (Tweet Paraphrasing), evaluated
with ROUGE-1 and ROUGE-L. Results are reported
as averages over three independent runs.}
\label{tab:detailed-lamp}
\end{table}

\begin{table}[!htbp]
\centering
\small
\setlength{\tabcolsep}{1.2pt}
\resizebox{\linewidth}{!}{%
\begin{tabular}{@{}>{\centering\arraybackslash}m{0.19\textwidth}>{\raggedright\arraybackslash}m{0.21\textwidth}*{6}{>{\centering\arraybackslash}m{0.09\textwidth}}@{}}
\toprule
\multirow{2}{*}{\textbf{Model}} & \multirow{2}{*}{\textbf{Method}} & \multicolumn{2}{c}{\makebox[0.184\textwidth][c]{\shortstack{Abstract\\Generation}}} & \multicolumn{2}{c}{\makebox[0.184\textwidth][c]{\shortstack{Product\\Review Generation}}} & \multicolumn{2}{c}{\makebox[0.184\textwidth][c]{\shortstack{Topic\\Generation}}} \\
\cmidrule(lr){3-4} \cmidrule(lr){5-6} \cmidrule(lr){7-8}
& & R-1 & R-L & R-1 & R-L & R-1 & R-L \\
\midrule
\multirow{6}{*}{DeepSeek-V4-Pro}
& Direct & 0.3040 & 0.1871 & 0.1408 & 0.0866 & 0.1926 & 0.0988 \\
& RAG & 0.3446 & 0.1978 & 0.2929 & 0.1406 & 0.2635 & 0.1229 \\
& RAG-SelfFactor & 0.3772 & 0.2178 & 0.3090 & 0.1508 & 0.1811 & 0.0998 \\
& RAG-CoPA & 0.3560 & 0.2022 & 0.3015 & 0.1426 & 0.2670 & 0.1248 \\
& Static PersonaFactor & 0.3613 & 0.2039 & 0.3124 & 0.1449 & 0.2685 & 0.1254 \\
& Dynamic PersonaFactor & 0.4173 & 0.2489 & 0.3624 & 0.1752 & 0.3139 & 0.1563 \\
\midrule
\multirow{6}{*}{Gemini-2.5-Flash}
& Direct & 0.3185 & 0.1913 & 0.1827 & 0.1031 & 0.1902 & 0.0983 \\
& RAG & 0.3561 & 0.2047 & 0.3200 & 0.1530 & 0.2603 & 0.1268 \\
& RAG-SelfFactor & 0.3332 & 0.1822 & 0.2696 & 0.1267 & 0.2040 & 0.1093 \\
& RAG-CoPA & 0.3647 & 0.2116 & 0.3651 & 0.1813 & 0.2886 & 0.1309 \\
& Static PersonaFactor & 0.3727 & 0.2160 & 0.3698 & 0.1841 & 0.2959 & 0.1366 \\
& Dynamic PersonaFactor & 0.4054 & 0.2365 & 0.4109 & 0.2002 & 0.3500 & 0.1787 \\
\midrule
\multirow{6}{*}{Qwen3-8B}
& Direct & 0.3126 & 0.1883 & 0.1851 & 0.1084 & 0.1475 & 0.0893 \\
& RAG & 0.3380 & 0.1944 & 0.2488 & 0.1315 & 0.1876 & 0.1052 \\
& RAG-SelfFactor & 0.3767 & 0.2078 & 0.2989 & 0.1516 & 0.2296 & 0.1187 \\
& RAG-CoPA & 0.3325 & 0.1910 & 0.2597 & 0.1351 & 0.2049 & 0.1111 \\
& Static PersonaFactor & 0.3341 & 0.1910 & 0.2626 & 0.1367 & 0.2189 & 0.1151 \\
& Dynamic PersonaFactor & 0.4024 & 0.2262 & 0.3299 & 0.1772 & 0.2808 & 0.1543 \\
\midrule
\multirow{6}{*}{Gemma-3-4B}
& Direct & 0.2907 & 0.1743 & 0.1554 & 0.0958 & 0.1553 & 0.0881 \\
& RAG & 0.2944 & 0.1764 & 0.2280 & 0.1246 & 0.1733 & 0.0975 \\
& RAG-SelfFactor & 0.2611 & 0.1572 & 0.1912 & 0.0997 & 0.1563 & 0.0898 \\
& RAG-CoPA & 0.2836 & 0.1725 & 0.2209 & 0.1216 & 0.1796 & 0.0995 \\
& Static PersonaFactor & 0.2884 & 0.1752 & 0.2304 & 0.1255 & 0.1849 & 0.1011 \\
& Dynamic PersonaFactor & 0.3576 & 0.2179 & 0.2806 & 0.1547 & 0.2446 & 0.1227 \\
\bottomrule
\end{tabular}%
}
\caption{Task-level results on LongLaMP under the user-setting split. The
tasks are Personalized Abstract Generation, Personalized Product Review
Generation, and Personalized Topic Generation; evaluation uses ROUGE-1 and
ROUGE-L. Results are reported as averages over three independent runs.}
\label{tab:detailed-longlamp}
\end{table}

\begin{table}[!htbp]
\centering
\small
\setlength{\tabcolsep}{3pt}
\resizebox{\linewidth}{!}{%
\begin{tabular}{@{}>{\centering\arraybackslash}m{0.20\textwidth}>{\raggedright\arraybackslash}m{0.40\textwidth}*{3}{>{\centering\arraybackslash}m{0.12\textwidth}}@{}}
\toprule
\multirow{2}{*}{\textbf{Model}} & \multirow{2}{*}{\textbf{Method}} & \shortstack{\scriptsize Art \&\\[-1pt] \scriptsize Entertainment} & \shortstack{\scriptsize Lifestyle \&\\[-1pt] \scriptsize Personal\\[-1pt] \scriptsize Development} & \shortstack{\scriptsize Society \&\\[-1pt] \scriptsize Culture} \\
\midrule
\multirow{6}{*}{DeepSeek-V4-Pro}
& Direct & 0.3764 & 0.4972 & 0.6009 \\
& RAG & 0.5306 & 0.6482 & 0.7720 \\
& RAG-SelfFactor & 0.5254 & 0.5983 & 0.6583 \\
& RAG-CoPA & 0.5886 & 0.7519 & 0.8182 \\
& Static PersonaFactor & 0.6027 & 0.7817 & 0.8409 \\
& Dynamic PersonaFactor & 0.6158 & 0.7970 & 0.8597 \\
\midrule
\multirow{6}{*}{Gemini-2.5-Flash}
& Direct & 0.3595 & 0.4331 & 0.5137 \\
& RAG & 0.4775 & 0.5639 & 0.6831 \\
& RAG-SelfFactor & 0.4211 & 0.5335 & 0.5989 \\
& RAG-CoPA & 0.4889 & 0.5719 & 0.6823 \\
& Static PersonaFactor & 0.4923 & 0.5840 & 0.6979 \\
& Dynamic PersonaFactor & 0.5199 & 0.6203 & 0.7021 \\
\midrule
\multirow{6}{*}{Qwen3-8B}
& Direct & 0.2573 & 0.3404 & 0.3837 \\
& RAG & 0.3259 & 0.4331 & 0.4888 \\
& RAG-SelfFactor & 0.2644 & 0.3823 & 0.4584 \\
& RAG-CoPA & 0.3420 & 0.4435 & 0.5055 \\
& Static PersonaFactor & 0.3443 & 0.4457 & 0.5084 \\
& Dynamic PersonaFactor & 0.3570 & 0.4542 & 0.5139 \\
\midrule
\multirow{6}{*}{Gemma-3-4B}
& Direct & 0.2693 & 0.4076 & 0.4516 \\
& RAG & 0.3568 & 0.4621 & 0.5161 \\
& RAG-SelfFactor & 0.3256 & 0.4317 & 0.4412 \\
& RAG-CoPA & 0.4075 & 0.5129 & 0.5740 \\
& Static PersonaFactor & 0.3991 & 0.5204 & 0.5638 \\
& Dynamic PersonaFactor & 0.4121 & 0.5358 & 0.5842 \\
\bottomrule
\end{tabular}%
}
\caption{Task-level results on LaMP-QA across Art and Entertainment, Lifestyle
and Personal Development, and Society and Culture. Scores are computed with
the benchmark's LLM-as-a-Judge protocol. Results are reported as averages over
three independent runs.}
\label{tab:detailed-lamp-qa}
\end{table}

\begin{table*}[t]
\centering
\small
\setlength{\tabcolsep}{6pt}
\begin{tabular}{@{}>{\centering\arraybackslash}m{0.20\textwidth}>{\raggedright\arraybackslash}m{0.40\textwidth}c@{}}
\toprule
Model & Method & PersonalLLM \\
\midrule
\multirow{6}{*}{DeepSeek-V4-Pro}
& Direct & -0.9013 \\
& RAG & 0.4049 \\
& RAG-SelfFactor & -0.3861 \\
& RAG-CoPA & 0.4642 \\
& Static PersonaFactor & 0.4795 \\
& Dynamic PersonaFactor & 0.4915 \\
\midrule
\multirow{6}{*}{Gemini-2.5-Flash}
& Direct & -0.4993 \\
& RAG & 0.2426 \\
& RAG-SelfFactor & -0.6971 \\
& RAG-CoPA & 0.2689 \\
& Static PersonaFactor & 0.2822 \\
& Dynamic PersonaFactor & 0.3022 \\
\midrule
\multirow{6}{*}{Qwen3-8B}
& Direct & -1.2800 \\
& RAG & -0.2778 \\
& RAG-SelfFactor & -0.3638 \\
& RAG-CoPA & -0.1901 \\
& Static PersonaFactor & -0.1876 \\
& Dynamic PersonaFactor & -0.1722 \\
\midrule
\multirow{6}{*}{Gemma-3-4B}
& Direct & -0.2998 \\
& RAG & -0.2414 \\
& RAG-SelfFactor & -0.2562 \\
& RAG-CoPA & -0.2012 \\
& Static PersonaFactor & -0.1977 \\
& Dynamic PersonaFactor & -0.1836 \\
\bottomrule
\end{tabular}
\caption{Task-level results on PersonalLLM. Scores are obtained from the
benchmark's preference reward models. Results are reported as averages over
three independent runs.}
\label{tab:detailed-personallm}
\end{table*}

\FloatBarrier
\subsection{\texorpdfstring{Effect of the Factor-selection Interface}{Effect of the Factor-selection Interface}}
\label{app:open-model-two-stage}

To assess the effect of the factor-selection interface independently
of the remaining pipeline, we instantiate Dynamic PersonaFactor with both
available selectors on the open-weight models, where both are feasible. The
two conditions use the same retrieved histories, PersonaFactor inventory,
generation prompts, and decoding settings, and differ only in how the
task-specific factor subset $\mathcal{S}_x$ is inferred.

\begin{table}[!htbp]
    \centering
    \small
    \fontfamily{ptm}\selectfont
    \setlength{\tabcolsep}{2.5pt}
    \renewcommand{\arraystretch}{1.10}
    \caption{Comparison of selector realizations within Dynamic
    PersonaFactor on open-weight models. The prompt-based selector directly
    predicts the factor subset, whereas the score-based selector uses
    task-relevance and generation-influence scores. LaMP and LongLaMP report macro-averaged
    ROUGE-1/ROUGE-L. LaMP-QA uses an LLM-as-a-Judge score, and PersonalLLM uses
    a reward-model score. Higher is better for all metrics, and the best result
    within each model is boldfaced. Results are averaged over three independent
    runs.}
    \label{tab:open-model-two-stage}
    \begin{tabular}{@{}llcccccc@{}}
        \toprule
        \multirow{2}{*}{\textbf{Model}}
            & \multirow{2}{*}{\textbf{Factor Selection}}
            & \multicolumn{2}{c}{\textbf{LaMP}}
            & \multicolumn{2}{c}{\textbf{LongLaMP}}
            & \textbf{LaMP-QA} & \textbf{PersonalLLM} \\
        \cmidrule(lr){3-4} \cmidrule(lr){5-6}
        \cmidrule(lr){7-7} \cmidrule(lr){8-8}
        & & R-1 & R-L & R-1 & R-L
            & {\scriptsize LLM Judge} & {\scriptsize Reward Model} \\
        \midrule
        \multirow{2}{*}{Qwen3-8B}
            & Prompt-based selector & 0.3199 & 0.2801 & 0.2612 & 0.1487
            & 0.4028 & -0.3255 \\
            & Score-based selector & \textbf{0.3336} & \textbf{0.2956}
            & \textbf{0.3377} & \textbf{0.1859} & \textbf{0.4417}
            & \textbf{-0.1722} \\
        \midrule
        \multirow{2}{*}{Gemma-3-4B}
            & Prompt-based selector & 0.2711 & 0.2378 & 0.2298 & 0.1299
            & 0.4672 & -0.2772 \\
            & Score-based selector & \textbf{0.2818} & \textbf{0.2490}
            & \textbf{0.2943} & \textbf{0.1651} & \textbf{0.5107}
            & \textbf{-0.1836} \\
        \bottomrule
    \end{tabular}
\end{table}
\FloatBarrier

When sequence-likelihood access is available, the score-based
selector consistently yields stronger results on the two open-weight models.
This indicates that direct likelihood scoring provides a more informative
selection signal in the evaluated settings. The prompt-based selector serves
a complementary access regime: it preserves the same PersonaFactor
representation, task-specific factor subset, and factor-conditioned generation
pipeline when sequence likelihoods are unavailable. The comparison therefore
reflects an access-performance trade-off between two selector realizations
within Dynamic PersonaFactor.

\subsection{Dynamic Selector Ablation on Gemma-3-4B}
\label{app:gemma-dynamic-selector-ablation}

We additionally evaluate the four dynamic selector variants on Gemma-3-4B
using the same LaMP-QA protocol and factor-selection settings as in
Section~\ref{sec:dynamic-selector-ablation}. Dynamic PersonaFactor achieves the highest score in all three
domains, while the single-signal and random-selection variants remain lower.

\begin{table}[!t]
    \centering
    \small
    \setlength{\tabcolsep}{4.5pt}
    \renewcommand{\arraystretch}{1.12}
    \caption{Dynamic selector ablation on Gemma-3-4B using LaMP-QA. Higher is
    better.}
    \label{tab:gemma-dynamic-selector-ablation}
    \begin{tabular}{@{}lccc@{}}
        \toprule
        \textbf{Selector}
            & \shortstack{Art \&\\Entertainment}
            & \shortstack{Lifestyle \&\\Personal Development}
            & \shortstack{Society \&\\Culture} \\
        \midrule
        Dynamic PersonaFactor & \textbf{0.4121} & \textbf{0.5358}
            & \textbf{0.5842} \\
        $R_{x,f}$-only & 0.3977 & 0.5086 & 0.5727 \\
        $G_{x,f}$-only & 0.4021 & 0.5198 & 0.5653 \\
        Random-$k$           & 0.4065 & 0.5156 & 0.5688 \\
        \bottomrule
    \end{tabular}
\end{table}

\FloatBarrier
\subsection{Effect of the Number of History Items on Gemma-3-4B}
\label{app:gemma-history-count-effect}

\begin{figure}[!htbp]
    \centering
    \includegraphics[width=\linewidth]{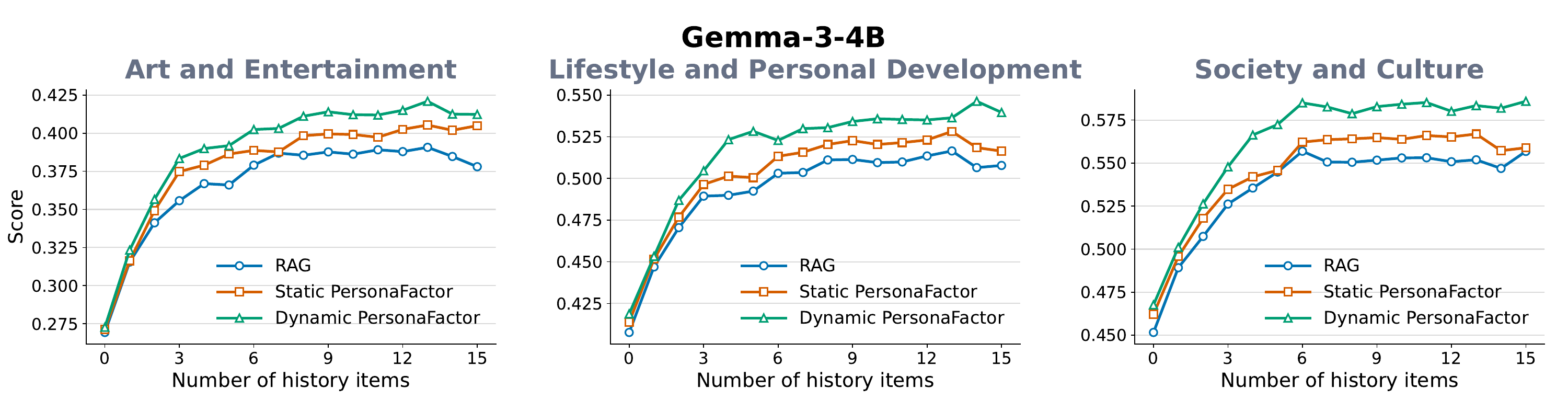}
    \caption{LaMP-QA performance as a function of the number of user-history
    items for Gemma-3-4B.}
    \label{fig:history-count-effect-gemma}
\end{figure}

We repeat the history-count analysis from
Section~\ref{sec:history-count-effect} on Gemma-3-4B. As shown in
Figure~\ref{fig:history-count-effect-gemma}, performance improves rapidly when
the first few history items are introduced and then generally saturates.
Dynamic PersonaFactor consistently outperforms RAG and Static PersonaFactor
across all three LaMP-QA domains and history budgets. The mild fluctuations at
larger history sizes further show that adding interactions alone does not
guarantee continued gains, supporting the benefit of selectively structuring
historical evidence on a second open-weight model.

\FloatBarrier

\subsection{Low-Dimensional Representation Analysis}
\label{app:low-dimensional-analysis}

For the analysis in Section~\ref{sec:low-dimensional-analysis}, we use ten
history items for LaMP and three for LongLaMP. At each transformer layer,
$h_B$, $h_{B,H}$, and $h_{B,F}$ are the final-prompt-token hidden states under
Direct, History, and Static PersonaFactor, respectively. A fixed rank-128
random projection $P\in\mathbb{R}^{128\times d}$, shared across conditions and
layers within each model, defines
\[
u_H=P(h_{B,H}-h_B),\qquad u_F=P(h_{B,F}-h_B),\qquad
\widehat{u}_c=\frac{u_c}{\lVert u_c\rVert_2},\quad c\in\{H,F\}.
\]
For each task, we split examples into 80\% training and 20\% held-out sets
while keeping each example's conditions and layers in the same split. We use
seed 42 for both the random projection and the training--held-out split. All
subspace bases are fitted only on training vectors.

We obtain the leading energy directions from the uncentered matrices of
unit-normalized effects, separately for History and PersonaFactor. For top-3
overlap, we pool vectors across layers $18$--$34$ for Qwen3-8B and
$17$--$32$ for Gemma-3-4B. All layer indices are one-based. For rank-8
overlap, we fit each layer separately. With orthonormal bases
$U_H,U_F\in\mathbb{R}^{128\times k}$, we compute
\[
S_k=\frac{1}{k}\left\lVert U_H^{\top}U_F\right\rVert_F^2
    =\frac{1}{k}\sum_{i=1}^{k}\sigma_i^2,\qquad k\in\{3,8\},
\]
where $\sigma_i$ are the singular values of $U_H^{\top}U_F$, or cosines of
the principal angles. The expected overlap of two random subspaces is
$k/128$: $0.0234$ for $k=3$ and $0.0625$ for $k=8$.

For the three-dimensional plots, normalized bisectors of the paired
canonical principal vectors define a shared display basis. We project
held-out unit vectors into this basis and smooth a $42^3$ histogram with a
Gaussian filter whose standard deviation is $\sigma=1.70$. Each condition's
region contains 80\% of its
density mass; green marks the regions' geometric intersection. This displayed
intersection is distinct from the subspace statistic $S_3$.

\begin{figure}[!htbp]
    \centering
    \includegraphics[width=\linewidth]{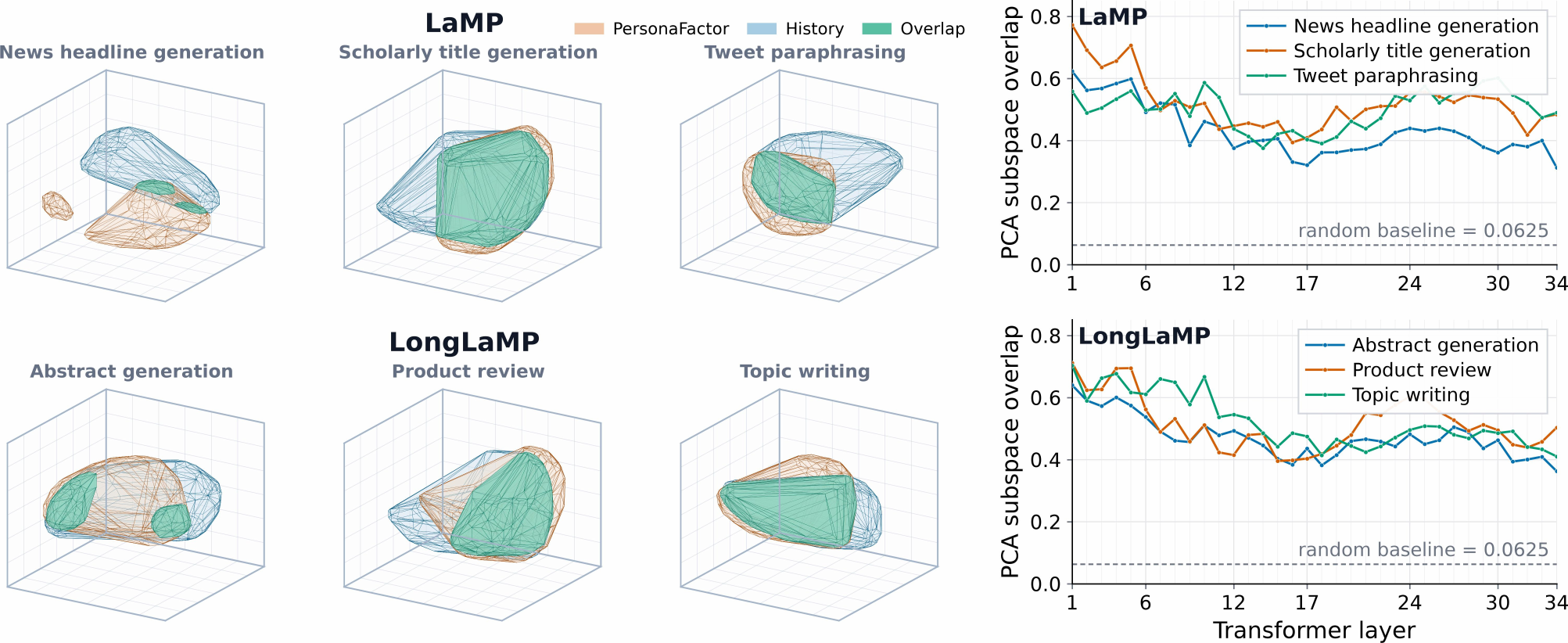}
    \caption{Low-dimensional overlap between History and PersonaFactor
    effects across LaMP and LongLaMP tasks and transformer layers for
    Gemma-3-4B.}
    \label{fig:low-dimensional-overlap-gemma}
\end{figure}

Figure~\ref{fig:low-dimensional-overlap-gemma} presents the corresponding
Gemma-3-4B analysis. Its task-level mean top-3 overlap is $0.370$, while its
task-level means for the layer-wise rank-8 overlap range from $0.430$ to
$0.523$. Both results substantially exceed their respective random-subspace
baselines, showing that the shared low-dimensional structure between History
and PersonaFactor also holds for Gemma-3-4B.

Table~\ref{tab:layer-wise-overlap-summary} reports the task-level means of
the layer-wise rank-8 overlap. The full-layer mean averages over all
transformer layers, while the middle-to-late-layer mean averages over layers
$18$--$34$ for Qwen3-8B and layers $17$--$32$ for Gemma-3-4B.

\begin{table*}[!t]
    \centering
    \small
    \begin{tabular}{lllc}
    \toprule
    Model & Dataset & Task & Top-3 overlap \\
    \midrule
    \multirow{6}{*}{Qwen3-8B}
        & \multirow{3}{*}{LaMP} & News headline generation & 0.180 \\
        & & Scholarly title generation & 0.181 \\
        & & Tweet paraphrasing & 0.293 \\
        & \multirow{3}{*}{LongLaMP} & Abstract generation & 0.272 \\
        & & Product review & 0.232 \\
        & & Topic writing & 0.257 \\
    \midrule
    \multirow{6}{*}{Gemma-3-4B}
        & \multirow{3}{*}{LaMP} & News headline generation & 0.197 \\
        & & Scholarly title generation & 0.470 \\
        & & Tweet paraphrasing & 0.439 \\
        & \multirow{3}{*}{LongLaMP} & Abstract generation & 0.393 \\
        & & Product review & 0.405 \\
        & & Topic writing & 0.317 \\
    \bottomrule
    \end{tabular}
    \caption{Task-level top-3 subspace overlap between History and
    PersonaFactor. The overlap is computed after aggregating the
    model-specific middle-to-late layer set.}
    \label{tab:task-level-overlap}
\end{table*}

\begin{table*}[!t]
    \centering
    \small
    \begin{tabular}{lllc c}
    \toprule
    Model & Dataset & Task & Full-layer mean & Middle-to-late-layer mean \\
    \midrule
    \multirow{6}{*}{Qwen3-8B}
        & \multirow{3}{*}{LaMP} & News headline generation & 0.352 & 0.333 \\
        & & Scholarly title generation & 0.437 & 0.434 \\
        & & Tweet paraphrasing & 0.395 & 0.368 \\
        & \multirow{3}{*}{LongLaMP} & Abstract generation & 0.380 & 0.356 \\
        & & Product review & 0.419 & 0.399 \\
        & & Topic writing & 0.437 & 0.428 \\
    \midrule
    \multirow{6}{*}{Gemma-3-4B}
        & \multirow{3}{*}{LaMP} & News headline generation & 0.430 & 0.391 \\
        & & Scholarly title generation & 0.521 & 0.503 \\
        & & Tweet paraphrasing & 0.499 & 0.507 \\
        & \multirow{3}{*}{LongLaMP} & Abstract generation & 0.471 & 0.446 \\
        & & Product review & 0.516 & 0.505 \\
        & & Topic writing & 0.523 & 0.469 \\
    \bottomrule
    \end{tabular}
    \caption{Task-level summary of layer-wise rank-8 subspace overlap
    between History and PersonaFactor.}
\label{tab:layer-wise-overlap-summary}
\end{table*}

\subsection{Combined History--PersonaFactor Effects}
\label{app:combined-effects-analysis}

This appendix formalizes the analysis in Section~\ref{sec:combined-effects-analysis}. For each example and
transformer layer, let $h_B$, $h_H$, $h_F$, and $h_{HF}$ denote the hidden state
of the final prompt token under the Direct, History-only, PersonaFactor-only,
and joint History+PersonaFactor conditions, respectively. All four conditions
use the same query, prompt template, decoding configuration, and history
budget; only the personalization inputs differ. Relative to Direct, we define

\[
u_H=h_H-h_B,\qquad u_F=h_F-h_B,\qquad
u_{HF}=h_{HF}-h_B.
\]

Under a purely additive model, the joint effect would equal $u_H+u_F$. We
therefore define the interaction residual as

\[
r_{HF}=u_{HF}-(u_H+u_F)=h_{HF}-h_H-h_F+h_B.
\]

To compare its magnitude across layers, models, and tasks, we report the
normalized residual

\[
\mathrm{NR}_{HF}=\frac{\lVert r_{HF}\rVert_2}
{\lVert u_H\rVert_2+\lVert u_F\rVert_2+\epsilon},
\]

where $\epsilon$ is a small positive constant. A small value indicates that
the joint shift is well explained by the two individual effects; a larger
value indicates stronger conditional interaction. We compute these quantities
for each observed layer and summarize them across the same six tasks used in
Section~\ref{sec:low-dimensional-analysis}. The resulting analysis separates reusable additive structure from
task-specific interaction without requiring access to model parameters or
output distributions.

\begin{figure}[!htbp]
    \centering
    \includegraphics[width=\linewidth]{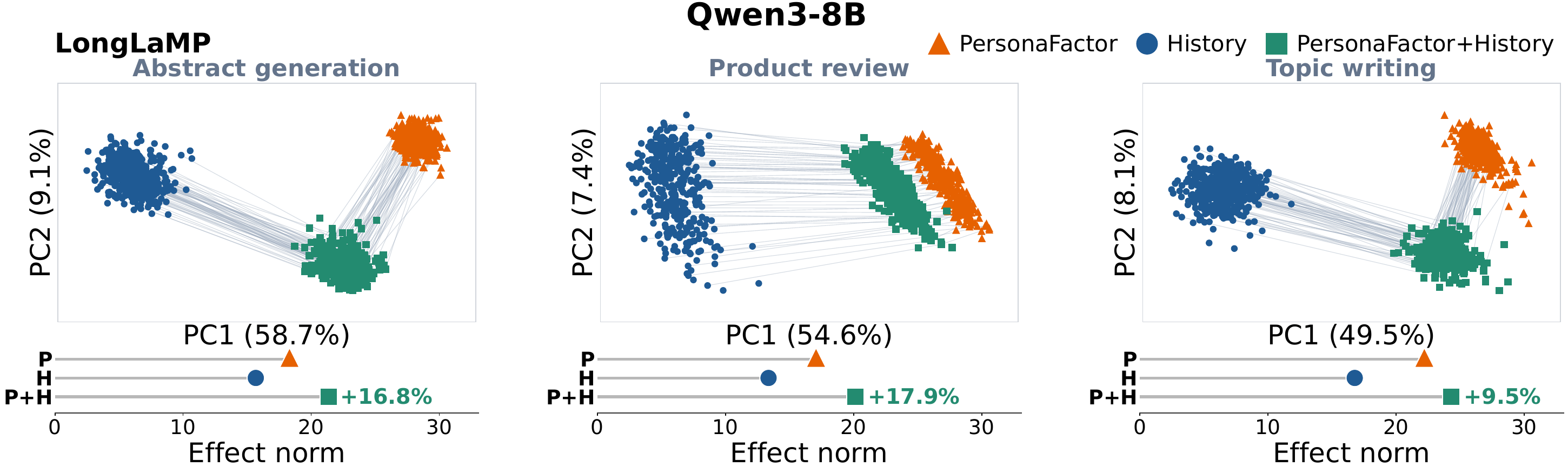}
    \caption{Combined activation effects of History and PersonaFactor for
    Qwen3-8B across LongLaMP tasks. Each task compares History, PersonaFactor,
    and their joint condition relative to the Direct baseline.}
    \label{fig:combined-effects-qwen-longlamp}
\end{figure}

%% file: 6_Limitations.tex
\section{Limitations}
\label{app:limitations}

Our study has several limitations. First, PersonaFactor construction relies on
language-model supervision for rationale extraction, mechanism naming, and
canonicalization, which may introduce systematic errors or model-dependent
biases. Second, the factor library is induced from a finite collection of
benchmarks and may not cover preference mechanisms that arise in other domains,
cultures, or interaction modalities. Third, dynamic factor selection requires
additional inference and therefore incurs computational overhead relative to
history retrieval alone. Finally, our experiments focus on text-based
personalization and do not explicitly model temporal changes in user
preferences. The representation analyses establish shared structure and
complementarity between History and PersonaFactor, but do not by themselves
establish a causal mechanism.